\documentclass{article}
\usepackage{ijcai23}

\usepackage{times}
\usepackage{soul}
\usepackage{url}
\usepackage[hidelinks]{hyperref}
\usepackage[utf8]{inputenc}
\usepackage[small]{caption}
\usepackage[symbol]{footmisc}
\renewcommand{\thefootnote}{\fnsymbol{footnote}}
\usepackage{graphicx}
\usepackage{amsmath}
\usepackage{amsthm}
\usepackage{booktabs}
\usepackage[switch]{lineno}

\usepackage[utf8]{inputenc}
\usepackage{xcolor}
\usepackage{graphicx}
\usepackage{multicol}
\usepackage{multirow}
\usepackage{array}
\usepackage{xspace}
\usepackage{enumitem}
\setlist[itemize]{leftmargin=*}
\setlist[enumerate]{leftmargin=*}
\usepackage{amsmath}
\usepackage{amsfonts,amssymb}
\usepackage[font=small,labelfont=bf]{caption}
\usepackage{tablefootnote}
\usepackage{comment}
\usepackage{hyperref}
\usepackage{braket}
\usepackage{bm}
\usepackage{subfigure}
\usepackage[ruled,linesnumbered]{algorithm2e}
\usepackage{algpseudocode}

\newcommand{\RL}{reinforcement learning }

\newcommand{\Imiss}{$\mathcal{N}_m\ $}
\newcommand{\Iobs}{$\mathcal{N}_o\ $}

\newtheorem{problem}{Problem}

\newcommand{\problemFont}{\mathit}
\newcommand{\problemSetFont}{\mathcal}
\newcommand{\states}{\problemSetFont{S}}
\newcommand{\observations}{\problemSetFont{S}}
\newcommand{\observation}{\problemFont{s}}
\newcommand{\observationsi}{\problemSetFont{\hat{S}}}
\newcommand{\observationi}{\problemFont{\hat{s}}}
\newcommand{\actions}{\problemSetFont{A}}

\newcommand{\reward}{\problemFont{r}}
\newcommand{\rewardi}{\problemFont{\hat{r}}}
\newcommand{\policy}{\problemFont{\pi}}

\newcommand{\decay}{\problemFont{\gamma}}

\newcommand{\methodFont}{\textsl}
\newcommand{\fixfix}{\methodFont{Fix-Fix}\xspace}
\newcommand{\idqnfix}{\methodFont{IDQN-Fix}\xspace}
\newcommand{\idqnidqn}{\methodFont{IDQN-IDQN}\xspace}
\newcommand{\sdqnsdqn}{\methodFont{SDQN-SDQN}\xspace}
\newcommand{\idqnmaxp}{\methodFont{IDQN-MaxP}\xspace}
\newcommand{\idqnn}{\methodFont{IDQN-Neighboring}\xspace}
\newcommand{\arxiv}[1] {{{{\color{black}{#1}}}}}
\renewcommand{\arraystretch}{1.5}

\title{Reinforcement Learning Approaches for Traffic Signal Control\\ under Missing Data}

\author{
Hao Mei$^1$
\and
Junxian Li$^2$\and
Bin Shi$^{2}$\And
Hua Wei\footnotemark[2]$^1$
\affiliations
$^1$  New Jersey Institute of Technology\\
$^2$  Xi’an Jiaotong University\\
\emails
\{hm467, hua.wei\}@njit.edu,
ljx201806@stu.xjtu.edu.cn,
shibin@xjtu.edu.cn
}

\begin{document}

\maketitle
\footnotetext[2]{Corresponding author.}

\renewcommand{\thefootnote}{\arabic{footnote}}
\begin{abstract}
The emergence of reinforcement learning (RL) methods in traffic signal control tasks has achieved better performance than conventional rule-based approaches. Most RL approaches require the observation of the environment for the agent to decide which action is optimal for a long-term reward. However, in real-world urban scenarios, missing observation of traffic states may frequently occur due to the lack of sensors, which makes existing RL methods inapplicable on road networks with missing observation. In this work, we aim to control the traffic signals in a real-world setting, where some of the intersections in the road network are not installed with sensors and thus with no direct observations around them. To the best of our knowledge, we are the first to use RL methods to tackle the traffic signal control problem in this real-world setting. Specifically, we propose two solutions: the first one imputes the traffic states to enable adaptive control, and the second one imputes both states and rewards to enable adaptive control and the training of RL agents. Through extensive experiments on both synthetic and real-world road network traffic, we reveal that our method outperforms conventional approaches and performs consistently with different missing rates. We also provide further investigations on how missing data influences the performance of our model. 
\end{abstract}

\section{Introduction}

Traffic congestion has been a challenge in modern society and adversely affects economic growth, environmental sustainability, and people's quality of life. For example, traffic congestion costs an estimated $\$87$ billion in lost productivity in the US alone~\cite{onathan2022Congestion}. Recently, reinforcement learning (RL) has shown superior performance over traditional transportation approaches in controlling traffic signals in dynamic traffic~\cite{arel2010reinforcement,el2013multiagent,oroojlooy2020attendlight,presslight19}. The biggest advantage of RL is that it directly learns to take adaptive actions in response to dynamic traffic by observing the states and feedback from the environment after previous actions.

Although a number of literature has focused on improving RL methods' performance in traffic signal control, RL cannot be directly deployed in the real world where accessible observations are sparse, i.e., some traffic states are missing~\cite{CHEN201973,mai2019inverse}. In most cities where sensors are only installed at certain intersections, intersections without sensors cannot utilize RL and usually use pre-timed traffic signal plans that cannot adapt to dynamic traffic. Similar situations could happen when installed sensors are not properly functioning, which will lead to missing observations in the collected traffic states~\cite{DUAN2016168} and the failure to deploy RL methods. \arxiv{Though there have been attempts such as using imitation learning to learn from the experience of human traffic engineers~\cite{li2020deep}, these methods need manual design and cannot be easily extended to new scenarios}. Thus the missing data issue still hinders not only the application of RL, but also the deployment of other adaptive control methods that require observing traffic states like MaxPressure~\cite{varaiya2013max}. %In real-world scenarios, accessible traffic observations are sparse, i.e., some traffic states are missing. 

In this paper, we investigate the traffic signal control problem under the real-world setting, where the traffic condition around certain intersections are never observed. %We investigate how data imputation could help the control. \aamas{To ensure that our approach converges, we discuss different kinds of RL agents, like Dueling DQN ~\cite{DBLP:journals/corr/WangFL15,DBLP:journals/corr/abs-2106-13174} and FRAP~\cite{DBLP:journals/corr/abs-1905-04722} on real-world setting.} 
To enable dynamic control over these intersections, we investigate how data imputation could help the control, especially how the imputation on state and reward can remedy the missing data challenge for RL methods. With imputed states, adaptive control methods from transportation could be utilized; with both imputed states and rewards, RL agents could be trained for unobserved intersections. Inspired by model-based RL, we also investigate to use the imaginary rollout with reward model for better performance.
The main contributions of this work are summarized as follows:

%In this paper, we proposed a two-step infrastructure called MissLight, to solve traffic signal control problems under missing data hence improving the viability of \RL in the real world. We specifically investigated intersection level missing data, which means all traffic data expect for phase information are missing at some specific intersections throughout the data collection. In this new approach, We first infer missing data with current state-of-the-art traffic data imputation methods. Then, we utilize the imputed data in \RL agents training and traffic signal control planning. We treat the traffic signal control under the missing data problem as a workflow with two separate components and optimize the whole process iteratively under the \RL framework. The main contributions of this work are summarized as follows:
\noindent$\bullet$ To the best of our knowledge, we are the first to adapt RL-based traffic signal control methods under the missing data scenario, hence improving \RL method's applicability under more realistic settings. We test different kinds of approaches to control the intersections without observations. We propose a two-step approach that firstly imputes the states and rewards to enable the second step of RL training. The proposed approach can achieve better performance than only training RL agents at fully observed intersections or training RL agents on all intersections using only observed data without imputation, and also outperforms using pre-timed control methods.
\\
\noindent$\bullet$ We investigate our methods under synthetic and real-world datasets with different missing rates and whether having neighboring unobserved intersections. The result shows our method is still effective in real-world scenarios. And the extensive studies on different missing rates and relationships of missing positions \arxiv{shows} our methods perform better than pre-timed methods. \arxiv{We also extend
our proposed methods for a highly heterogeneous dataset and justified their effectiveness, enhancing their applicability in real life.}

\section{Related Work}

\vspace{1mm}
\noindent \textbf{Traffic signal control methods.} Optimizing traffic signal control to alleviate traffic congestion has been a challenge in the transportation field for a long time. Different approaches have been extensively studied, including rule-based methods~\cite{hunt1981scoot,sims1980sydney,varaiya2013max} and RL-based methods~\cite{arel2010reinforcement,presslight19,wei2018intellilight} to optimize vehicle travel time or delay. Most of these studies,  \arxiv{for example, IDQN method~\cite{wei2018intellilight}}, have significantly improved compared to pre-designed time control methods. %IDQN in ~\cite{wei2018intellilight} deploys an agent at each intersection, and each agent learns its policy based on its own experiences. This method successfully optimizes traffic signal control and is easy to transform to other intersections with the same road structures and signal phases. 
However, to the best of our knowledge, there is no existing work on dealing with the unobserved intersections in dynamic traffic signal control methods.% In this paper, we take the IDQN method as the baseline model and build our proposed methods to solve the traffic signal control problem under the missing data scenario.

\vspace{1mm}
\noindent \textbf{Traffic data imputation.}
In real-world scenarios, full observation is not always accessible. %And at the same time, RL-based and most rule-based traffic control methods need observation at each intersection during execution to decide which phase to take. 
An effective way to deal with the missing observations %and make RL and other rule-based methods work properly with missing data 
is data imputation, i.e., to infer the missing data to complete traffic observations. %There have been many investigations with various approaches to alleviating the sparsity problem in traffic data collection. 
Earlier studies typically use historical data collected at each location to predict the values at missing positions of the same site~\cite{6953131,ZHONG2004139}. %However, these approaches only use data collected at the same positions and neglect the spatial dependencies of the existing traffic road networks. 
Recently, neural network-based methods have been proven effective and extended to be used in the traffic data imputation task~\cite{6894591}. These methods could be categorized into Recurrent Neural Networks (RNNs)~\cite{CUI2020102674,Yao_Wu_Ke_Tang_Jia_Lu_Gong_Ye_Li_2018}, Graph Neural Network (GNN)~\cite{wang2022multi} and Generative Adversarial Networks (GANs)~\cite{zhang2021missing}. However, all the methods mentioned above also need the observation of all intersections to train models, while in reality, it is hard to fulfill. Store-and-forward method (SFM)~\cite{ABOUDOLAS2009163} is another approach to model traffic state transition and is often used as the base model traffic simulation.

\vspace{1mm}
\noindent \textbf{Model-based reinforcement learning.}
Model-based reinforcement learning (MBRL) methods utilize predictive models of the environment on the immediate reward or transition to provide imaginary samples for RL~\cite{https://doi.org/10.48550/arxiv.2206.09328}. %Compared to the model-free RL, it has higher data efficiency and can be extended to partially observable environments~\cite{ma2022conservative, vertes2019neurally}. 
In MBRL with a reward model, an agent learns to predict the immediate reward of taking action at a certain state, and in this paper, we borrow this idea to train a reward model. 
In MBRL with a transition model, an agent usually has the direct observation of its own surrounding states, and a transition model is used to simulate the \emph{next states} from current observations. Unlike MBRL with transition models, in this paper, some agents do not have observations of their own surroundings, where imputation methods are utilized to infer the \emph{current states} (rather than next states) for those agents. 

\section{Preliminaries}
\label{sec:prelim}
% In this section, we take the basic problem definition used in the multi-intersection traffic signal control ~\cite{}-(presslight) and extend it into the missing data scenario that are frequently encountered in the real world. In the following parts, we introduce some basic definitions used in the transportation domain, then define the new problem as a variant of the original traffic signal control problem. In the last part, we introduce a frequently used method in the transportation domain to impute missing traffic data.

% \subsection{Problem: traffic signal control under missing data}

In this section, we take the basic problem definition used in the multi-intersection traffic signal control~\cite{presslight19} and extend it into the missing data scenario frequently encountered in the real world. An agent controls each intersection in the system. Given that only part of the agents can have their local observation of the total system condition as their state, we would like to proactively decide for all the intersections in the system which phases they should change to so as to minimize the average queue length on the lanes around the intersections.  Specifically, the problem is characterized by the following major components $\braket{\observations,\observationsi, \actions, \reward,\Pi,\decay}$:

$\bullet$ 
Observed state space $\observations$ and imputed state space $\observationsi$. We assume that the system consists of a set of intersections $\mathcal{N} = \mathcal{N}_o \cup \mathcal{N}_m$, where $\mathcal{N}_o$ is the set of intersections where the agent can observe part of the system as its state $\observation \in \observations$, and $\mathcal{N}_m$ is the set of intersections where the agent cannot observe the system. We follow setting from past works~\cite{wei2019colight,wu2021dynstgat,huang2021modellight}, and define
% In this work, we define 
$\observation_t^j$ for agent $j \in \mathcal{N}_o$ at time $t$, which consists of its current phase (which direction is in green light) and the number of vehicles on each lane at time $t$. Later we will introduce unobserved agent $k \in \mathcal{N}_m$, and how we can infer its state $\observationi_t^k \in \observationsi$ at time $t$.

$\bullet$ 
Set of actions $\actions$. In the traffic signal control problem, at time $t$, an agent $i$ would choose an action $a_t^i$ from its candidate action set  $\actions_i$ as a decision for the next $\Delta t$ period of time. Here, we 
 take acyclic control method, in which  each intersection would choose a phase $p$ as its action $a_t^i$ from its pre-defined phase set, indicating that from time $t$ to $t+\Delta t$, this intersection would be in phase $p$.

% $\bullet$
% Transition probability $\transition$. Given the system state $\state^t$ and corresponding joint actions $\actionVec^t$ of agents at time $t$, the system arrives at the next state $\state^{t+1}$ according to the state transition probability $\transition(\state^{t+1}| \state^t,\actionVec^t): \states \times \actions_1 \times \dots \times \actions_N \rightarrow \Omega(\states)$, where $\Omega(\states)$ denotes the space of state distributions.

$\bullet$
Reward $\reward$. Each agent $i$ obtains an immediate reward $\reward_t^i$ from the environment at time $t$ by a reward function $\states \times \actions_1 \times \dots \times \actions_N \rightarrow \mathbb{R}$. In this paper, we want to minimize the travel time for all vehicles in the system, which is hard to optimize directly. Therefore, we define the reward for intersection $i$ as $\reward_t^i = -\Sigma_l u_t^{i,l}$ where $u_t^{i,l}$ is the queue length on the approaching lane $l$ at time t. Specifically, we denote $\reward_t^j$ as the observed reward for agent $j \in \mathcal{N}_o$ at time $t$, and the inferred reward $\rewardi_t^k$ as the reward for agent $k \in \mathcal{N}_m$ at time $t$.

$\bullet$
Policy set $\Pi$ and discount factor $\decay$. Intuitively, the joint actions have long-term effects on the system, so we want to minimize the expected queue length of each intersection in each episode. Specifically, at time $t$, each agent chooses an action following a certain policy $\policy: \observations \rightarrow \actions$.
% $\observations\times \actions\rightarrow \policy$.

An RL agent follows policy $\policy_\theta \in \Pi$ parameterized by $\theta$, aiming to maximize its total reward $G_t^i=\Sigma_{\tau=t}^T \decay^{\tau-t}\reward_t^i$, where $T$ is total time steps of an episode and $\decay\in[0,1]$ differentiates the rewards in terms of temporal proximity. Other rule-based agents are denoted as $\policy_\varnothing \in \Pi$.

\begin{problem}[Traffic signal control under missing data]
Given a road network where only part of the intersections is observed with $\observations$, the goal of this paper is to find a better $\Pi$, no matter whether it consists of $\policy_\theta$, $\policy_\varnothing$ or mixed policies of previous two kinds, that can minimize the average travel time of all vehicles.
\end{problem}

In the RL framework, training and execution are two decoupled phases: 
(1) During \emph{execution}, an agent takes actions based on its policy $\policy$ to roll out trajectories and evaluate their performances. For policies that take the current state as input, the agent can execute adaptive actions as long as the input states are available. For observed intersections $\mathcal{N}_o$, the input states could directly be observed state $\observation \in \observations$; for unobserved intersections $\mathcal{N}_m$, the input states could be inferred $\observationi \in \observationsi$ using data imputation. (2) In the \emph{training} phase, agents explore the environment, store experiences in the replay buffer, and update their policies to maximize their long-term rewards. The experiences usually consist of state $\observation_t$, reward $\reward_t$, action $a_t$, and next state $\observation_{t+1}$. Different from \emph{execution} phase, which only requires the input states, the \emph{training} phase of RL  requires reward information. For unobserved intersections $\mathcal{N}_m$, the $\rewardi_t$ could also be inferred with data imputation on the reward. % As a result, only the current observation $S_t$ is needed for the agent's execution. 
Later in Sec.~\ref{sec: methods}, we will introduce how the missing data in the training and execution phase would influence the design of methods to tackle the traffic signal control problem.

		\section{Methods under Missing Data}
\label{sec: methods}

\begin{figure*}[ht]
\centering
\includegraphics[width=0.95\linewidth]{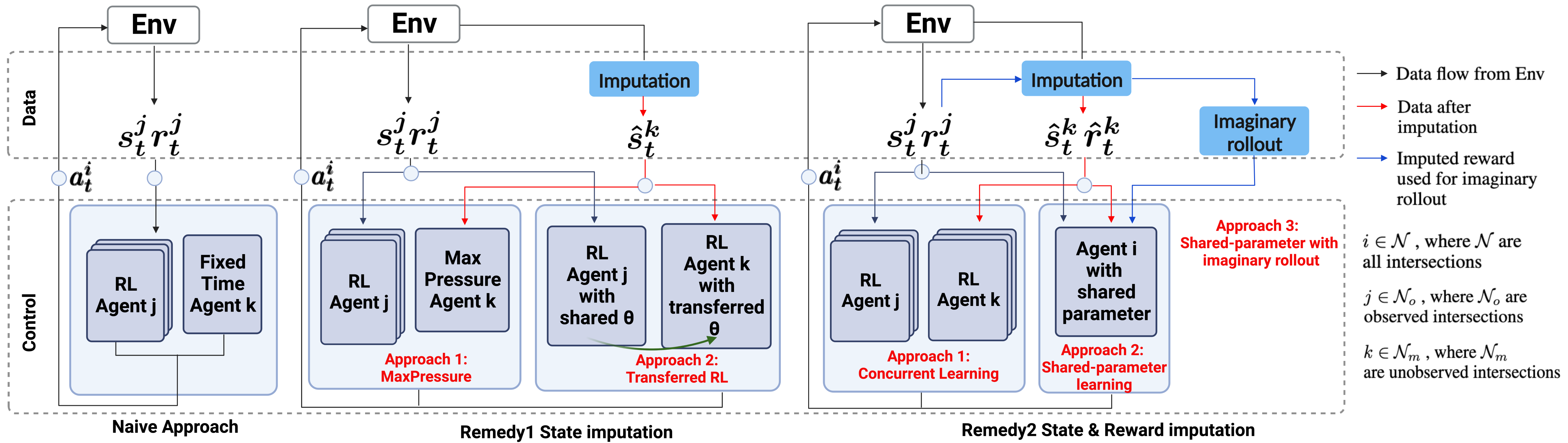}
\vspace{-3mm}
\caption{Overall framework of our proposed approaches. Red lines represent how the imputed data flows into unobserved intersections. And blue lines show how reward is imputed and used for imaginary rollout. In Remedy 1, we only impute states at \Imiss and enable agents at these intersections to take adaptive actions; In Remedy 2, we impute states and rewards together for training agents. \emph{(a) Remedy 1.1}: RL agents at \Iobs take $s_t^j, r_t^j$ for training their policies, and take $s_t^j$ during the execution phase. And MaxPressure agents at \Imiss take imputed $\hat{s}_t^k$ together with neighboring intersections' $s_t^j$ during training and execution phases. \emph{(b) Remedy 1.2}: RL agents take $s_t^j, r_t^j$, and train based on these observable experiences. And at the $\mathcal{N}_m$, agents are transferred from \Iobs and take imputed $\hat{s}_t^k$ for taking actions during the execution phase. \emph{(c) Remedy 2.1}: agents at \Iobs and \Imiss take $s_t^j,r_t^j \text{ pairs or } \hat{s}_t^k, \hat{r}_t^k$ pairs respectively and optimize their own policies during training. During execution, agents at \Iobs take $s_t^j$ from the environment, and agents at \Imiss take imputed $\hat{s}_t^k$ for execution. Different from Remedy 2.1, in \emph{(d) Remedy 2.2} and \emph{(e) Remedy 2.3}, all agents share the same policy, while \emph{Remedy 2.3} has an additional imaginary rollout step.}
\vspace{-3mm}
\label{fig:Overall framework}
\end{figure*}

Adaptive control methods like MaxPressure and RL-based methods dynamically adjust traffic signals based on real-time traffic state observations, which have been proven to work well on traffic signal control tasks. %For example, independent deep Q-network (IDQN)~\cite{wei2018intellilight} in which each agent independently learns a deep Q-Network to estimate reward, has improved greatly compared to the pre-timed control method. %For example, Max Pressure~\cite{varaiya2013max}, which changes phases based on its own and neighboring intersections' observations, performs better than pre-timed control methods. The RL-based methods, another dynamic traffic signal control method, also achieved significant results in recent years. 
However, in the real-world scenario, these adaptive control methods cannot work properly at intersections $\mathcal{N}_m$ where real-time observations are missing.
To adapt dynamic control methods to real-world, we explore the conventional approach and propose two effective imputation approaches to handle the failure of adaptive control at  $\mathcal{N}_m$. %\aamas{In addition, at $\mathcal{N}_o$, we include two more RL agents to enhance the adaption to real-world environment of our method.}
%To stress data missing problems in real-world scenarios, we introduce imputation methods to complete data first. The RL framework can then use the imputed data to solve traffic signal control problems.
The overall frameworks are shown in Figure~\ref{fig:Overall framework}.

%In this section, we present two approaches, state imputation, and state and reward imputation, to recover missing data for dynamically controlling traffic signals and compare the result with the conventional adaptive one. We pick a rule-based method, max pressure, and a Q-learning method, DQN~\cite{wei2018intellilight}, which has been proven effective in traffic signal control problems but with complete observation as the archetype. Then, we build these two imputation approaches upon it and investigate how they can help alleviate the missing data problem in the traffic signal control task under real-world scenarios.

\subsection{Conventional Approaches}
\label{sec: conventional approach}
%  \vspace{1mm}
% \noindent \textit{\textbf{Pre-timed control.}} 
Under the missing data scenario, there are three direct approaches for traffic signal control: (1) pre-timed control, which sets fixed timings for all intersections; (2) \label{sec: conventional 2}the mixed control of RL and pre-timed agents, which uses RL agents only at observed intersections $\mathcal{N}_o$ and deploys pre-timed agents at unobserved intersections $\mathcal{N}_m$; %The RL agents can reach local optimality at $\mathcal{N}_o$ while keeping functionality at $\mathcal{N}_m$.
% \vspace{1mm}
% \noindent \textit{\textbf{Combining RL and pre-fixed control.}}
%As we will show in the experiments, the performances of RL agents on $\mathcal{N}_o$ are better than pre-timed agents. 
%Since the agents at $\mathcal{N}_m$ still take non-dynamic execution and are not optimized, the overall performance of the conventional approach is significantly reduced compared to optimizing all agents at every intersection $\mathcal{N}$ ideally.
%  \ijcai{
% \vspace{1mm}
% \noindent \textit{\textbf{Neighboring RL control.}} 
(3) neighboring RL control, where agents at unobserved intersections $\mathcal{N}_m$ concatenate states from neighboring observed intersections as their own state and accumulate rewards from their neighboring observed intersections as their reward. This approach follows the general solution to the Partially Observable Markov Decision Process (POMDP) and assumes the traffic condition from observed neighboring intersections could reflect the unobserved traffic condition. This assumption might not hold when the number of missing intersections increases or the traffic is dynamic and complex. % Using neighboring intersections information can help RL agents work on unobserved intersections without imputation step. However, the new state and reward cannot reflect the real information, thus causing performance degeneration.

%We sample unobserved intersections at the most ideal setting in which only one intersection is unobserved and all its neighboring intersections are observed.

\subsection{Remedy 1: Imputation over Unobserved States}
To enable $\mathcal{N}_m$ with dynamic control during execution, a natural solution is to impute unobserved states $\hat{s}_t^k$ at \Imiss for control methods. After the imputation of the states at unobserved intersections, dynamic control methods can be applied.
%by using states $s_t^j$ at observed intersections $\mathcal{N}_o$. }

\paragraph{Imputation.}
\label{sec: Imputation}
% Since we permanently lose state information at unobserved intersections, simply applying the supervised learning method to train a model and recover unobserved states is not applicable in our setting. SFM model is a rule-based method that is often used in past traffic signal control design avenues. In this work, since we lack knowledge of the condition of each unobserved intersection $j$, we model current state $s_{t}^{j,l} = \Sigma_i \Sigma_{l_{in}} s_{t-1}^{i,l_{in}} / n$, where $i$ is the observed up-streaming intersections of $j$ and $s_{t-1}^{i,l_{in}}$ is the state at incoming lane $l_{in}$ at time $t-1$ and $n$ is the number of all incoming lanes to lane $l$. While all up-streaming intersections are unobserved, we use the average on all observed intersections to impute intersection $j$. Besides the ruled-based SFM model, we also use a Spatial-temporal graph modeling method, GraphWN, to impute the state following the past work~\cite{lei2022modeling,wu2019graph}.
% Details are shown in Table~\ref{tab: imputation and accuracy}.
Since the state information at unobserved intersections is totally missing, it is inapplicable to train a model on data collected from unobserved intersections and recover unobserved states. Therefore, we need to pretrain a state imputation model that will be shared by all the unobserved intersections and apply it during the training of RL. Intuitively, vehicles currently on each lane are aggregated from its up-streaming connected lanes in the previous time step. Given the states of neighboring intersections $\mathcal{V}^k$ of intersection $k \in \mathcal{N}_m$, the state imputation at $k$ can be formally defined as follows:
\begin{equation}
\label{eq:impute-s}
\hat{s}_t^k = f(\mathcal{V}^k_{t-1})
\end{equation}
where $f$ could be any state imputation model. In this paper, we investigate two pre-trained models, a rule-based Store-and-Forward model (SFM) and a neural network model. Their detailed descriptions can be found in Sec.~\ref{sec: experiment imputation setting}.

\paragraph{Control.}
After imputation, we investigate two control approaches that can function during execution.

\vspace{1mm}
\noindent$\bullet$ \textit{Approach 1: Adaptive control methods in transportation.} Adaptive control methods in transportation usually require observation of the surrounding traffic conditions to decide the action for traffic signals. Without imputation, these methods cannot be applied directly. In this paper, following~\cite{presslight19,chen2020toward}, we use one widely used adaptive control method, MaxPressure~\cite{varaiya2013max}, to control traffic signals for unobserved intersections after imputation.

\vspace{1mm}
\noindent$\bullet$ \textit{Approach 2: Transferred RL models.}
\label{sec: Controlling with trans DQN}
Another method is to enable RL-based control at missing intersections by training an RL policy $\policy_\theta$ at observed intersections \Iobs and later transferring to \Imiss during execution. Since all agents share the same policy, we refer to this model-sharing agent as SDQN for later use. During execution, agent $j \in \mathcal{N}_o$ can directly use the states $s^j_t$ observed from the environment to take action $\policy_\theta(a^j_t| \observation^j_t)$. For agent $k \in \mathcal{N}_m$, it first imputes states $\observationi^k_t$ and then takes action $\policy_\theta(a^k_t| \observationi^k_t)$ based on the imputed states $\observationi^k_t$. In this solution, we use $(s^j_t, r^j_t, a^j_t, s^j_{t+1})$ from all the observed intersections $j \in \mathcal{N}_o$ as experiences to train an RL model shared by all intersections. This approach can significantly improve sample efficiency, and all training samples can reflect the true state of the environment. However, since the agent is only trained on the experiences from observed intersections $\mathcal{N}_o$, it might not be able to cope with unexplored situations at $\mathcal{N}_m$, which could result in a loss of generality based on the agent's policy. 

\subsection{Remedy 2: Imputation over Unobserved States and Rewards}
To enable agents to learn from experiences on unobserved intersections $\mathcal{N}_m$, it is necessary to impute both state and reward for unobserved intersections $k \in \mathcal{N}_m$. After getting both the imputed state $\observationi_t^k$ and inferred reward $\hat{r}_t^k$ at $\mathcal{N}_m$, we can train agents with these imputed experiences $(\observationi_t^k, \hat{r}_t^k, a_t^k, \observationi_{t+1}^k)$.

\paragraph{Imputation.}
% We choose SFM to impute state  $\observationi_t^k$ and a pre-trained reward imputation model to infer reward $\rewardi_t^k$ for $\mathcal{N}_m$. In the pre-training phase, we first run simulations to collect $(s^j_t, r^j_t, a^j_t)$ as training samples from observed intersections $j \in \mathcal{N}_o$, upon which we train the reward imputation model with $s^j_{t}$ and $a^j_t$ as input to predict $\hat{r}^j_t$.
The process of state imputation is the same as described in Sec.~\ref{sec: Imputation}. 
% For reward imputation, we use a neural network to infer $\rewardi_t^k$ for $\mathcal{N}_m$ and pre-train it before RL training starts. In the pre-training phase, we first run simulations with conventional control approach to collect $(s^j_t, r^j_t, a^j_t)$ as training samples from observed intersections $j \in \mathcal{N}_o$, upon which we train the reward imputation model with $s^j_{t}$ and $a^j_t$ as input to predict $\hat{r}^j_t$}. During the RL training phase, the original {$s_t^j, r_t^j$} returned from the environment will first pass through \ijcai{state imputation} model and get the recovered data $\hat{s}_t$. Then we take $\hat{s}^k_{t}$ and $a^k_t$, where $k \in \mathcal{N}_m$ as input into the pre-trained reward imputation model to impute $\hat{r}_t^k$. Combining  $r_t^j$ from $j \in \mathcal{N}_o$, experiences at all intersections $(\hat{s}_t, \hat{r}_t, a_t, \hat{s}_{t+1})$ are available now.
For reward imputation, we use a neural network to infer $\rewardi_t^k$ for $\mathcal{N}_m$ with state and action as input and pre-train it before RL training starts. In the pre-training phase, we first run with a conventional control approach to collect $(s^j_t, r^j_t, a^j_t)$ as training samples from observed intersections $j \in \mathcal{N}_o$, upon which we train the reward imputation model $g_{\psi}$ with MSE Loss:
\begin{equation}
\label{eq:impute-r-loss}
\begin{split}
    % & \hat{r}_j = g_{\psi}(s_j,a_j), \\
    & \mathcal{L}(\psi) = \frac{1}{n} \sum_n {( g_{\psi}(s^j,a^j) - r^j)^2}
\end{split}
\end{equation}
% with $s^j_{t}$ and $a^j_t$ as input to predict $\hat{r}^j_t$.
During the RL training phase, the original $s_t^j$ at \Iobs returned from the environment will first pass through state imputation model and get the recovered data $\hat{s}^k_t$ at \Imiss. The imputed $\hat{s}_t^k$ combining $a_t^k$ will be fed into $g_{\psi}$ which could be described as:
\begin{equation}
\label{eq:impute-r}
\hat{r}^k_t =  g_{\psi}(\hat{s}^k_t, a_t^k)
\end{equation}
Combining  $r_t^j$ from $j \in \mathcal{N}_o$, experiences at all intersections are now available.

\paragraph{Control.}
After state and reward imputation, the problem of traffic signal control under the missing data could be transformed into the regular traffic signal control problem. In the following, we investigate three approaches:

\vspace{1mm}
\noindent$\bullet$ \textit{Approach 1: Concurrent learning.}
\label{sec: concurrent learning}
In concurrent learning, each agent has its own policy and learns from its own experiences.  We adopt this imputation over state and reward approach to enable the training of RL method. We use experiences returned from the environment to train RL agents at \Iobs and imputed experiences 
%$(\hat{s}_t^k, \hat{r}_t^k, a_t^k, \hat{s}^k_{t+1})$ 
to train agents at $\mathcal{N}_m$. This training approach concurrently trains agents over all intersections and potentially makes each agent achieve its local optimality if the evaluation metric could converge at the training end. The concurrent training process could be problematic when the imputation at missing intersections is inaccurate, and training on such imputed experiences can bring additional uncertainties and make it hard to get stable RL models. 

\vspace{1mm}
\noindent$\bullet$ \textit{Approach 2: Parameter sharing.}
To improve the sample efficiency and reduce the instability during training, we investigate the shared-parameter learning approach as Sec.~\ref{sec: Controlling with trans DQN} did. During training, we collect observed experiences $(s^j_t, r^j_t, a^j_t, s^j_{t+1})$ from observed intersections $j \in \mathcal{N}_o$ and use the imputation models to impute $(\hat{s}_t^k, \hat{r}_t^k)$ and get the imputed experiences $(\hat{s}_t^k, \hat{r}_t^k, a_t^k, \hat{s}^k_{t+1})$ for intersections $k \in \mathcal{N}_m$. Then a shared RL policy is trained with both observed and imputed experiences.
During execution, the trained RL policy is shared by all the intersections. This parameter-sharing approach aims to expose the shared agent to the experiences from both $\mathcal{N}_o$ and $\mathcal{N}_m$ and make policy stable and easy to converge including experiences collected from heterogeneous structure datasets~\cite{https://doi.org/10.48550/arxiv.2005.13625,DBLP:journals/corr/abs-1905-04722}.

\vspace{1mm}
\noindent$\bullet$ \textit{Approach 3: Parameter sharing with the imaginary rollout.}
In all imputation approaches, we use a rule-based SFM and pre-trained neural network to impute states or states and rewards. However, the sample distribution shifting caused by different policies could be detrimental to the performance of the pre-trained model~\cite{chen2019information}. Thus we combine the model-based \RL (MBRL) with the reward model and train a shared policy in the Dyna-Q style framework~\cite{sutton1991dyna,zhao2020dynamic}. 

In this approach, the shared-parameters agent updates the Q function with both observed experiences $(s^j_t, r^j_t, a^j_t, s^j_{t+1})$ from observed intersections and imputed experiences from state and reward imputation models. At each simulation step, the reward imputation model $g_{\psi}$ infers $\hat{r}^j, j \in$ \Iobs, which will be used in training $g_{\psi}$ by calculating the loss between $\hat{r}^j$ and $r^j$ returned from the environment with Eq.~\eqref{eq:impute-r}. Each round of imaginary rollout samples a batch of $(s_c^j, a_c^j, s'^j_c)$ and $(\hat{s}_c^k, a_c^k, \hat{s}'^k_c)$, where $ k \in$ \Imiss and $ j \in$ \Iobs. For $ k \in$ \Imiss, the updated reward imputation model will infer the new $\hat{r}_c^j, \hat{r}_c^k$ to apply additional updates the Q function:
\begin{equation}
\label{eq:q-learning}
Q_\theta(s_c^i, a_c^i) := \hat{r}_c^i + \gamma \max_{a'^i} Q_\theta(s'^i_c, a'^i)
\end{equation}
where,  $\hat{r}_c^i \in \{\hat{r}_c^j, \hat{r}_c^k\}$.
% $\hat{r}_c^i \in \{\hat{r}_c^j, \hat{r}_c^k\},(s_c^i, a_c^i, {s'}_c^i) \in \{(s_c^j,a_c^j,{s'}_c^j), (\hat{s}_c^k,a_c^k,{\hat{s}'}_c^k)\}$
Details are shown in Algorithm~\ref{algo:2.3}.

\begin{algorithm}[htb]
\DontPrintSemicolon
\caption{Algorithm for Remedy 2.3 - \sdqnsdqn (model-based) with imaginary rollout}
\label{algo:2.3}

\KwIn{Observed intersections $j\in$ \Iobs, unobserved intersections $k\in$ \Imiss, pre-trained reward function $g_\psi(s,a)$, initial $Q_\theta(s,a)$,  state imputation function $f(\mathcal{V}^k_{t-1})$}
\KwOut{$Q_\theta$, $g_\psi$}

	\For {e = 1,2, \dots}
	{
    	Reset simulator environment \;
    	\For {t = 1,2 , \dots}
    	{
        	 Use states from observed intersections to impute $\hat{s}^k$ with Eq.~\eqref{eq:impute-s} \;
        	 Take actions with $a^j=\arg\max Q_\theta(s^j, a^j)$, and $a^k=\arg \max Q_\theta(\hat{s}^k, a^k)$ \;
             Infer $\hat{r}_k$ with $g_\psi(\hat{s}^k, a^k)$, and $\hat{r}_j$ with $g_\psi(s^j, a^j)$ \;
             Record $r^j$ returned from environment \;
        	 Update $Q_\theta(s,a)$ with $(s^j, r^j, a^j, s'^j)$ and $(\hat{s}^k, \hat{r}^k, a^k,\hat{s}'^k)$ using Eq.~\eqref{eq:q-learning}\;
    	
        	\textbf{\textit{\# Reward function update step}} \;
        	 Optimizing $g_\psi(s,a)$ with Eq.~\eqref{eq:impute-r-loss} \;
        	\textbf{\textit{\# Imaginary rollout step}} \;
            \For {c = 1,2 , \dots}
            {
             Randomly sample experiences, including  imputated state-action pairs $(\hat{s}^k, a^k, \hat{s}'^k)$ from \Imiss and observed experiences $(s^j, a^j, s'^j)$ \;
             Infer $\hat{r}^j$ with $g_\psi(s^j, a^j)$, $\hat{r}^k$ with $g_\psi(\hat{s}^k, a^k)$ \;
             Update $Q_\theta(s,a)$ with $(s^j, \hat{r}^j, a^j, s'^j)$ and $(\hat{s}^k, \hat{r}^k, a^k, \hat{s}'^k)$ \;
            
            }
    	}
	}
\end{algorithm}

\section{Experiments}

\begin{table*}[t]

\centering
\caption{Overall performance of two imputation approaches and two baseline approaches w.r.t. the average travel time. The lower, the better. The \underline{\textbf{best}} and \underline{\normalfont second best} performance are highlighted. `-' means this method does not converge after 100 epochs of training.}
\label{tab:overall performance}
\resizebox{\textwidth}{!}{%
\begin{tabular}{|c|c|cccccccc|} \hline
\multirow{2}{*}[-1.5ex]\textbf{Dataset} &
 \multirow{2}{*}{\textbf{ Missing rate}} &
  \multicolumn{8}{c|}{\textbf{Method}} \\ \cline{3-10} 
 &
   &
  \multicolumn{1}{c|}{\textbf{\fixfix}} &
  \multicolumn{1}{c|}{\textbf{\idqnn}} &
  \multicolumn{1}{c|}{\textbf{\idqnfix}} &
  \multicolumn{1}{c|}{\textbf{\idqnmaxp}} &
  \multicolumn{1}{c|}{\textbf{\sdqnsdqn (transferred)}} &
  \multicolumn{1}{c|}{\textbf{\idqnidqn}}
  & \multicolumn{1}{c|}{\textbf{\sdqnsdqn (all)}} &
  \textbf{\sdqnsdqn (model-based)} \\ \hline
\multirow{4}{*}{\textbf{$D_{HZ}$}} &
  \textbf{6.25\%} &
  \multicolumn{1}{c|}{\multirow{4}{*}{609.13}} &
  \multicolumn{1}{c|}{433.67$\pm$26.75} &
  \multicolumn{1}{c|}{337.07$\pm$4.54} &
  \multicolumn{1}{c|}{334.41$\pm$2.42} &
  \multicolumn{1}{c|}{331.16$\pm$2.28} &
  \multicolumn{1}{c|}{424.81$\pm$15.36} &
  \multicolumn{1}{c|}{ \underline{ 330.85}$\pm$2.61} & \underline{\textbf{ 330.23}}$\pm$1.04  \\ \cline{2-2} \cline{4-10}
 &
  \textbf{12.5\%} & \multicolumn{1}{c|}{}
   &
  \multicolumn{1}{c|}{-} &
  \multicolumn{1}{c|}{362.89$\pm$6.03} &
  \multicolumn{1}{c|}{339.71$\pm$1.86} &
  \multicolumn{1}{c|}{\underline{330.84}$\pm$1.85} &
  \multicolumn{1}{c|}{497.21$\pm$59.43} &
  \multicolumn{1}{c|}{\underline{\textbf{329.11}}$\pm$0.30}  &  331.35$\pm$1.63  \\ \cline{2-2} \cline{4-10} 
 &
  \textbf{18.75\%} &
  \multicolumn{1}{c|}{} &
  \multicolumn{1}{c|}{-} &
  \multicolumn{1}{c|}{370.18$\pm$2.58} &
  \multicolumn{1}{c|}{342.57$\pm$1.82} &
  \multicolumn{1}{c|}{\underline{ 332.20}$\pm$4.55   } &
  \multicolumn{1}{c|}{537.85$\pm$56.67 } & \multicolumn{1}{c|}{\underline{\textbf{330.28}}$\pm$1.99}
    & 358.55$\pm$35.78 \\   \cline{2-2} \cline{4-10} 
 &
  \textbf{25\%} &
  \multicolumn{1}{c|}{} &
  \multicolumn{1}{c|}{-} &
  \multicolumn{1}{c|}{396.36$\pm$1.79} &
  \multicolumn{1}{c|}{382.93$\pm$5.60} &
  \multicolumn{1}{c|}{\underline{331.57}$\pm$1.81 } &
  \multicolumn{1}{c|}{653.09$\pm$71.44 } & \multicolumn{1}{c|}{ 333.87$\pm$3.06} & \underline{\textbf{330.51}}$\pm$1.48 \\ \hline
\multirow{4}{*}{\textbf{$D_{SYN}$}} &
  \textbf{6.25\%} &
  \multicolumn{1}{c|}{\multirow{4}{*}{713.69}} &
  \multicolumn{1}{c|}{767.54$\pm$14.63} &
  \multicolumn{1}{c|}{640.68$\pm$27.11} &
  \multicolumn{1}{c|}{577.92$\pm$31.58} &
  \multicolumn{1}{c|}{ \underline{350.85}$\pm$7.51} &
  \multicolumn{1}{c|}{683.41$\pm$95.21} & \multicolumn{1}{c|}{ 368.76$\pm$4.43 } & \underline{\textbf{325.64}}$\pm$13.31  \\ \cline{2-2} \cline{4-10} 
 &
  \textbf{12.5\%} &
  \multicolumn{1}{c|}{} &
  \multicolumn{1}{c|}{-} &
  \multicolumn{1}{c|}{600.45$\pm$15.53} &
  \multicolumn{1}{c|}{699.26$\pm$54.01} &
  \multicolumn{1}{c|}{\underline{399.03}$\pm$15.75} &
  \multicolumn{1}{c|}{727.64$\pm$83.78} & \multicolumn{1}{c|}{440.68$\pm$43.54 } & \underline{\textbf{361.07}}$\pm$9.64\\ \cline{2-2} \cline{4-10} 
 &
  \textbf{18.75\%} &
  \multicolumn{1}{c|}{} &
  \multicolumn{1}{c|}{-} &
  \multicolumn{1}{c|}{637.50$\pm$32.93 } &
  \multicolumn{1}{c|}{673.56$\pm$42.76} &
  \multicolumn{1}{c|}{808.25$\pm$52.81 } &
  \multicolumn{1}{c|}{794.89 ± 51.09} & \multicolumn{1}{c|}{\underline{584.03}$\pm$37.95 }
 & \underline{\textbf{568.21}}$\pm$7.29  \\ \cline{2-2} \cline{4-10} 
 &
  \textbf{25\%} &
  \multicolumn{1}{c|}{} &
  \multicolumn{1}{c|}{-} &
  \multicolumn{1}{c|}{ 574.01$\pm$18.42} &
  \multicolumn{1}{c|}{719.52$\pm$28.91  } &
  \multicolumn{1}{c|}{660.59$\pm$17.09 } &
  \multicolumn{1}{c|}{ 877.44$\pm$101.36 } & \multicolumn{1}{c|}{\underline{\textbf{538.42}}$\pm$32.09}
   & \underline{540.13}$\pm$20.17\\ \hline
\multirow{4}{*}{\textbf{$D_{NY}$}} &
  \textbf{6.25\%} &
  \multicolumn{1}{c|}{\multirow{4}{*}{1099.67}} &
  \multicolumn{1}{c|}{519.95$\pm$259.09} &
  \multicolumn{1}{c|}{286.43$\pm$124.59} &
  \multicolumn{1}{c|}{334.41$\pm$2.42} &
  \multicolumn{1}{c|}{200.72$\pm$9.1 } &
  \multicolumn{1}{c|}{279.25$\pm$40.34} &
  \multicolumn{1}{c|}{\underline{\textbf{191.06}}$\pm$1.21} & \underline{192.53}$\pm$4.26\\ \cline{2-2} \cline{4-10} 
 &
  \textbf{12.5\%} &
  \multicolumn{1}{c|}{} &
  \multicolumn{1}{c|}{-} &
  \multicolumn{1}{c|}{726.68$\pm$163.72} &
  \multicolumn{1}{c|}{502.56$\pm$285.56 } &
  \multicolumn{1}{c|}{\underline{215.54}$\pm$3.39} &
  \multicolumn{1}{c|}{ 336.9$\pm$26.28} &
  \multicolumn{1}{c|}{513.08$\pm$273.82} & \underline{\textbf{210.4}}$\pm$5.36 \\ \cline{2-2} \cline{4-10} 
 &
  \textbf{18.75\%} &
  \multicolumn{1}{c|}{} &
  \multicolumn{1}{c|}{-} &
  \multicolumn{1}{c|}{913.48$\pm$31.77 } &
  \multicolumn{1}{c|}{820.91$\pm$82.97} &
  \multicolumn{1}{c|}{240.98$\pm$33.12 } &
  \multicolumn{1}{c|}{415.31$\pm$83.85} &
  \multicolumn{1}{c|}{\underline{228.39}$\pm$2.73} & \underline{\textbf{220.49}}$\pm$1.06 \\ \cline{2-2} \cline{4-10} 
 &
  \textbf{25\%} &
  \multicolumn{1}{c|}{} &
  \multicolumn{1}{c|}{-} &
  \multicolumn{1}{c|}{1012.91$\pm$44.25} &
  \multicolumn{1}{c|}{ 1218.71$\pm$32.86} &
  \multicolumn{1}{c|}{\underline{414.79}$\pm$147.33} &
  \multicolumn{1}{c|}{1331.60$\pm$58.94 } &
  \multicolumn{1}{c|}{507.67$\pm$181.38}  & \underline{\textbf{316.56}}$\pm$37.97 \\ \hline
\end{tabular}%
}
\vspace{-3mm}
\end{table*}

\subsection{Experimental Setup}
\paragraph{Datasets}
We testify our two imputation approaches on traffic signal control tasks under missing data on a synthetic dataset and two real-world datasets. 

\vspace{1mm}
\noindent$\bullet$ \textbf{$D_{SYN}$:} This is a synthetic dataset generated by CityFlow~\cite{zhang2019cityflow}, an open-source microscopic traffic simulator. The traffic road network is $4\times 4$ grid structured, and traffic flow is randomly generated following Gaussian distribution.

\vspace{1mm}
\noindent$\bullet$ \textbf{$D_{HZ}$:} This is a public traffic dataset that recorded a $4 \times 4$ network at Hangzhou in 2016. All the dataset is collected from surveillance cameras nearby. 
%The data contains every vehicle's position and speed at each second. And it also contains the trajectory of each vehicle within the road network.

\vspace{1mm}
\noindent$\bullet$ \textbf{$D_{NY}$:} This is a public traffic dataset collected in New York City within $16\times 3$ intersections.
%The contents  are the same as described in the $D_{HZ}$ dataset. 

Three datasets contain every vehicle's position and speed at each second and the trajectory within the road network. \arxiv{And all three datasets are publicly available~\footnote{\url{https://traffic-signal-control.github.io/}}}

\paragraph{Implementation}
In this section, we introduce the details of \RL, state and reward imputation models during implementation.
% To pre-train the reward imputation model, we use a four-layer feed-forward neural network and simulate 100 epochs to collect the training data with traffic signals controlled by \idqnfix. The training samples are collected from observed intersections and divided into 80\% and 20\% for training and testing. In the RL framework, we train agents for 100 epochs and take the average travel time for agents' performance evaluation.

\vspace{1mm}
\noindent$\bullet$ \textit{RL settings.}
We follow the past work~\cite{wei2019colight,wu2021dynstgat,huang2021modellight} to set up the RL environment, and details on the state, reward, and action definition can be found in Sec.~\ref{sec:prelim}. We take exploration rate $\epsilon = 0.1$, discount factor $\gamma = 0.95$, minimum exploration rate $\epsilon_{min} = 0.01$, exploration decay rate $\epsilon_{decay} = 0.995$, and model learning rate $r = 0.0001$.

\vspace{1mm}
\noindent$\bullet$ \textit{State imputation model.}
\label{sec: experiment imputation setting}
SFM model is a rule-based method often used in past traffic signal control design avenues. In this work, we model current state as:
$
    f(\mathcal{V}^k_{t-1}) = \frac{1}{|\mathcal{V}^k_{t-1}|}{\Sigma_{l} s_{t-1}^{l}}
$,
where $l \in \mathcal{V}^k_{t-1}$ and $|\mathcal{V}^k_{t-1}|$ is the number of k's neighboring intersections.
In this paper, we also investigated a neural network model with Spatial-temporal Graph Neural Network, GraphWN~\cite{wu2019graph}, to impute the states but found out that SFM model has generally more stable performances. Their experiment results can be found in Appendix~\ref{tab: imputation model}.
% Details are shown in Table~\ref{tab: imputation and accuracy}.
% where $l \in \mathcal{V}_k$ and $n = |\mathcal{V}_k|$ 
% In this work, since we lack knowledge of the condition of each unobserved intersection $j$, we model current state $s_{t}^{j,l} = \Sigma_i \Sigma_{l_{in}} s_{t-1}^{i,l_{in}} / n$, where $i$ is the observed up-streaming intersections of $j$ and $s_{t-1}^{i,l_{in}}$ is the state at incoming lane $l_{in}$ at time $t-1$ and $n$ is the number of all incoming lanes to lane $l$. While all up-streaming intersections are unobserved, we use the average on all observed intersections to impute intersection $j$. Besides the ruled-based SFM model, we also use a Spatial-temporal graph modeling method, GraphWN, to impute the state following the past work~\cite{lei2022modeling,wu2019graph}.
% Details are shown in Table~\ref{tab: imputation and accuracy}.

\vspace{1mm}
\noindent$\bullet$ \textit{Reward imputation model.} To pre-train the reward imputation model, we use a four-layer feed-forward neural network and simulate 100 epochs to collect the training data with traffic signals controlled by the conventional approach 2 described in Sec.~\ref{sec: conventional 2}. The training samples are collected from observed intersections and divided into 80\% and 20\% for training and testing. In the RL framework, we train agents for 100 epochs and take the average travel time for agents' performance evaluation.

\paragraph{Compared methods} To describe different control methods without misunderstanding, we use the kind of agents at observed and unobserved intersections to denote these methods. For example, in~\idqnfix, the first term represents that $\mathcal{N}_o$ uses IDQN~\cite{wei2018intellilight}, and the second term represents that $\mathcal{N}_m$ uses fixed timing. %The name and control method mapping can be found in Table~\ref{Table: name of approaches}

% \begin{table}[ht]
% \centering
% \caption{Mapping between different approaches described in ~\ref{sec: methods} and its reference name in experiments. }
% \label{Table: name of approaches}
% \begin{tabular}{|l|c|}
% \hline
% \textbf{Approach name}                    & \textbf{Reference name (Obsv - Miss)} \\ \hline
% Conventional 1: FixedTime agents                   & Fix - Fix                             \\ \hline
% Conventional 2: local optimization with observed & IDQN - Fix                            \\ \hline
% Remedy 1.1: Controlling with MaxPressure        & IDQN - MaxP                           \\ \hline
% Remedy 1.2: Controlling with transfered DQN     & SDQN - SDQN (transferred)                    \\ \hline
% Remedy 2.1: concurent learning               & IDQN - IDQN                           \\ \hline
% Remedy 2.2: parameter-sharing learning       & SDQN - SDQN (all)                     \\ \hline
% \end{tabular}
% \end{table}

% \paragraph{Baseline Control Methods}
% To compare the performances of different approaches in MissLight's workflow, we take two naive approaches to solving traffic signal control tasks with missing data as baseline models. The first one is FixedTime control methods which assign FixedTime agents at every intersection. And the second one is local optimization at observed intersections.

\vspace{0mm}
\noindent$\bullet$  \textit{Conventional 1: \fixfix.} This is a ruled-based method with fixed timings for all phases. We use Webster's method~\cite{koonce2008traffic} to calculate the fixed timing and fine-tune it with a grid search to ensure the fixed time method had its best results.

\vspace{0mm}
\noindent$\bullet$ \textit{Conventional 2: \idqnfix.} In this method, intersections in \Iobs use their own model trained by Deep Q-Learning (DQN)~\cite{wei2018intellilight} and intersections in $\mathcal{N}_m$ use fine-tuned fixed timings. 
% \end{itemize}

\vspace{0mm}
\noindent$\bullet$ \textit{Conventional 3: \idqnn.} This is a method where both \Imiss and \Iobs use IDQN. At \Iobs, agents take in state and reward from the environment, and at $\mathcal{N}_m$, agents take states and rewards from neighboring intersections. Unobserved neighboring intersections are zero-padded.

% \paragraph{Proposed Methods}

% From the result in ~\ref{tab:overall performance}, we can see optimizing IDQN agents at \Iobs can reduce the average travel time significantly, but as the number of unobserved intersections increases, the performance gets worse and downgrades to the FixedTime control approach as all intersections observations are missing.

% \subsection{Experiments on Proposed Remedies}
% In this section, we perform experiments on different imputation approaches and their corresponding control strategies. 

% \begin{itemize}
\vspace{0mm}
\noindent$\bullet$  \textit{Remedy 1.1: \idqnmaxp.} In this method, intersections in $\mathcal{N}_o$ uses the same IDQN agents as~\idqnfix. For intersections in $\mathcal{N}_m$, a ruled-based control approach MaxPressure~\cite{varaiya2013max} is used after the imputation of $\hat{s}_t^k$. Different from the conventional methods, this method has a pre-defined SFM model for state imputation, which is shared by all the intersections in $\mathcal{N}_m$. %the observed states $s_t^j$ could be used directly to recover data at \Imiss and hereafter used in Maxpressure control. 
 
\vspace{0mm}
\noindent$\bullet$  \textit{Remedy 1.2: \sdqnsdqn (transferred).} Similar to \idqnmaxp, this method also imputes the states with SFM model. Different from \idqnmaxp, all the agents share one policy which is trained by collecting data from intersections in $\mathcal{N}_o$ and then transferred to intersections in \Imiss. 

% This control method can be fulfilled by imputing the $\hat{s}_t^k$ at $\mathcal{N}_m$. SDQN agents sharing one policy are deployed at all intersections. During training phase, only observed $(a_t^j, r_t^j, a_t^j, s_{t+1}^j) \text{ where } j \in \mathcal{N}_o$ experiences could be used to train SDQN agents. During the execution phase, the learned policy will be transferred to \Imiss and all agents choose actions with the same policy.
    
\vspace{0mm}
\noindent$\bullet$  \textit{Remedy 2.1: \idqnidqn.} Unlike Remedy 1, in addition to state imputation model, this method has a pretrained reward imputation model shared by the intersections in $\mathcal{N}_m$.
Each intersection has its individual RL policy to control the actions trained from the observed data (for intersections in $\mathcal{N}_o$) or the imputed data (for intersections in $\mathcal{N}_m$).

%\noindent$\bullet$  \textit{Remedy 2.1: \idqnidqn.} \ In this control method, both imputed states $\hat{s}_t^k $ and inferred rewards $\hat{r}_t^k $ are required. During training, each agent takes one intersection and learns from the experiences at this intersection. At the $\mathcal{N}_o$, experiences are returned from the environment, and at $\mathcal{N}_m$, experiences are inferred from two sequential imputation models. During execution, agents directly take in states from the environment at \Iobs and pick actions according to their individual policies. And for agents at $\mathcal{N}_m$, the states are first imputed from observed states, then put into the policies to determine agents' actions. 

\begin{figure*}[thb]
\centering
  \begin{tabular}{ccc}
  \includegraphics[width=.31\linewidth]{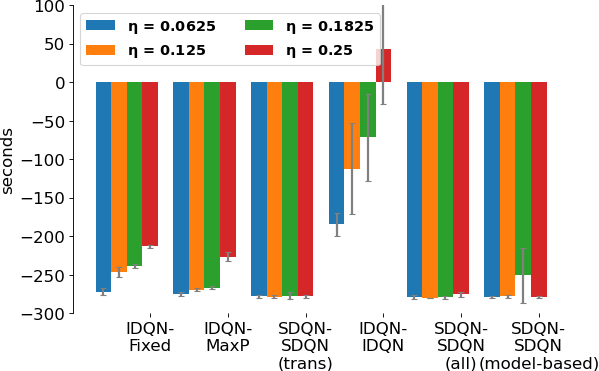} &
  \includegraphics[width=.31\linewidth]{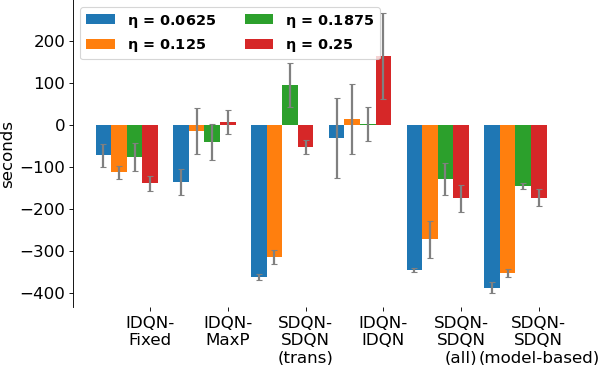} &
  \includegraphics[width=.31\linewidth]{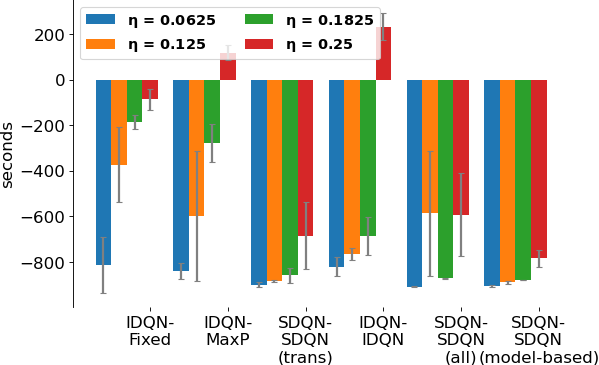}\\
  \vspace{-1mm}
  (a) $D_{HZ}$ & (b) $D_{SYN}$  & (c) $D_{NY}$  \\
  \end{tabular}
    \vspace{-2mm}
 \caption{The decrease of average travel time for six control approaches over \fixfix control method (conventional 1) with missing data at non-neighboring intersections. The more negative the value is, the better. Each column in one group uses the same control method. $\eta$ represents the missing rate of intersections in the road network. All dynamic control methods achieve better performance than \fixfix control. 
 \idqnmaxp, \sdqnsdqn (transferred),  \sdqnsdqn (all), and \sdqnsdqn (model-based) outperform \idqnfix (conventional 2) method on all three datasets.
 % \idqnmaxp, \sdqnsdqn (transferred), and \sdqnsdqn (all) outperform \idqnfix (conventional 2) method on all three datasets.
 }
    \label{fig:non-neigbboring}
 \vspace{-2mm}
\end{figure*}

% \begin{figure*}[tbh]
% \centering
%   \begin{tabular}{ccc}
%   \includegraphics[width=.27\linewidth]{figures/drop_hz.png} &
%   \includegraphics[width=.27\linewidth]{figures/drop_syn.png} &
%   \includegraphics[width=.27\linewidth]{figures/drop_mh.png}\\

%   \vspace{-1mm}
%   (a) $D_{HZ}$ & (b) $D_{SYN}$  & (c) $D_{NY}$  \\
%   \end{tabular}
%     \vspace{-2mm}
%  \caption{The decrease of average travel time for six control approaches over \fixfix control method (conventional 1) with missing data at non-neighboring and neighboring intersections. The more negative the value is, the better. Each column in one group uses the same control method and transparent ones are missing at neighboring intersections. Missing data at neighboring intersections introduces a significant challenge to our two steps imputation and control approaches.}
%     \label{fig:neigbboring}
% \end{figure*}

\begin{figure}[tb]
\centering
  \includegraphics[width=0.9\linewidth]{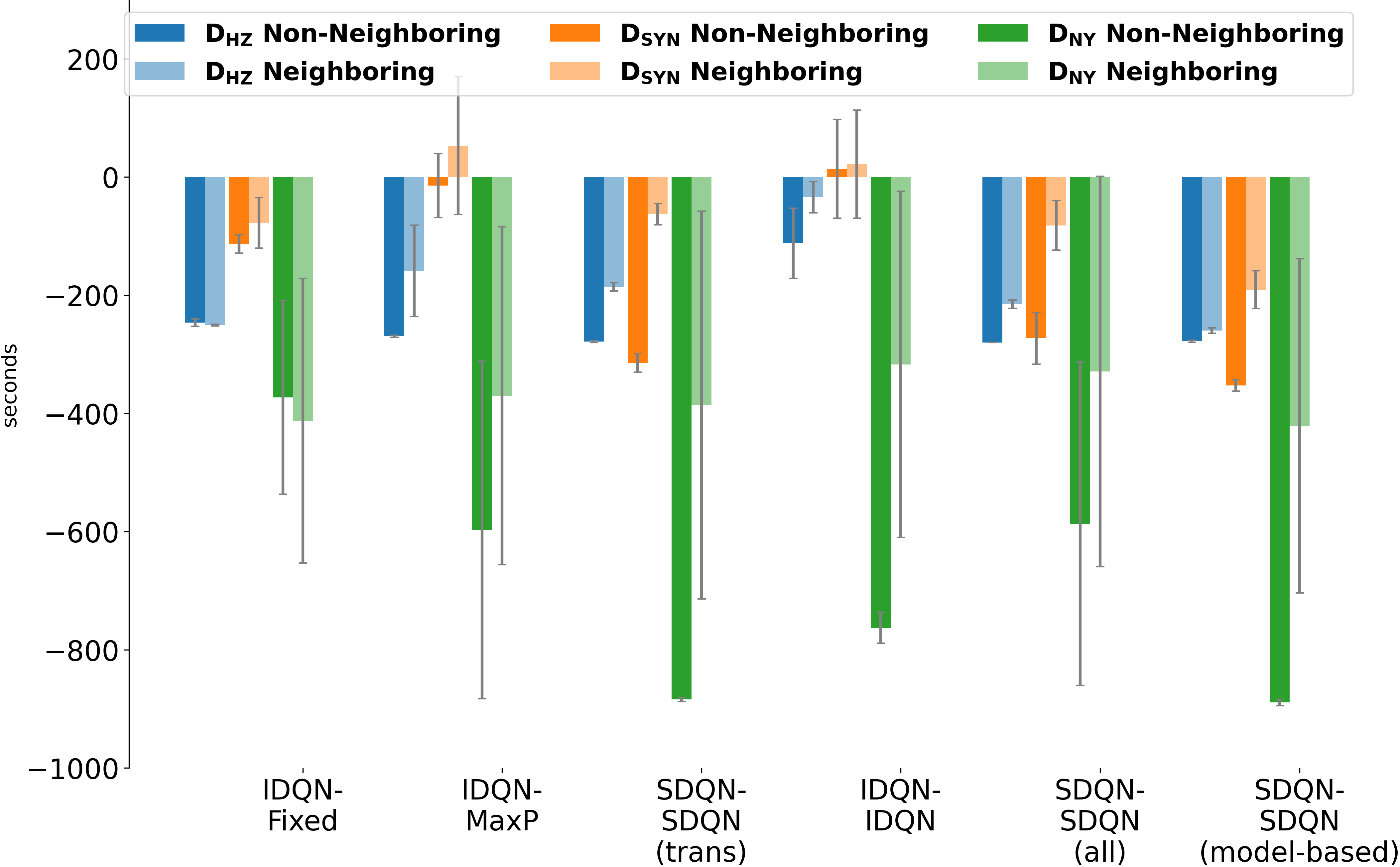}
  \vspace{-2mm}
 \caption{The decrease of average travel time for six control approaches over \fixfix (conventional 1) with missing data at non-neighboring and neighboring intersections under all three datasets. The more negative the value is, the better. Each group on X-axis represents a method. Within each group, three colors represent three datasets, where light- and dark-colored bars indicate scenarios with and without adjacent unobserved intersections, respectively. Our proposed method outperforms the \fixfix method in most cases.}
    \label{fig:neigbboring}
 \vspace{-2mm}
\end{figure}

\noindent$\bullet$  \textit{Remedy 2.2: \sdqnsdqn (all).} Similar to \idqnidqn, this method also has a state imputation model and a reward imputation model, while it only trains one shared policy using the observed data from \Iobs and imputed data from \Imiss.

%While this method uses the same parameter-sharing DQN method to train one shared agent, the fundamental of this approach is not the same as \sdqnsdqn (transferred). During training, the SDQN agent learns from experiences at all intersections, thus needing both $\hat{s}_t^k$ and $\hat{r}_t^k$. Shared DQN takes training samples from all intersections to learn one policy during training. The execution procedure is the same as \idqnidqn.

\vspace{0mm}
\noindent$\bullet$  \textit{Remedy 2.3: \sdqnsdqn (model-based).} This method integrates \sdqnsdqn (all) approaches into MBRL framework. During training, the SDQN agents learn from experiences at all intersections and, at the same time, the reward inference model predicts reward at $\mathcal{N}_m$. Different from \sdqnsdqn (all), the reward model is updated upon new experiments returned from the environment; it also has an imaginary rollout phase, during which SDQN agents learn from both states $s_t^j$ at \Iobs and imputed states $\hat{s}_t^k $ at \Imiss and inferred rewards $\hat{r}_t^i $ from the updated reward model.

% \paragraph{Experiment settings and metrics}
% This section introduces the experimental settings and the naming method of different control approaches in our work. 
% To pre-train the reward imputation model, we use a four-layer feed-forward neural network and simulate 100 epochs to collect the training data with traffic signals controlled by \idqnfix. The training samples are collected from observed intersections and divided into 80\% and 20\% for training and testing. In the RL framework, we train agents for 100 epochs and take the average travel time for agents' performance evaluation.

\subsection{Overall Performance}

% From the result in ~\ref{tab:overall performance}, we can see optimizing IDQN agents at \Iobs can reduce the average travel time significantly, but as the number of unobserved intersections increases, the performance gets worse and downgrades to the FixedTime control approach as all intersections observations are missing.

% In both three datasets, we can see that SDQN-SDQN (all), SDQN-SDQN (transferred), and IDQN-MaxP approaches achieve better performances than naive IDQN-Fix approaches, which proves the effectiveness of our two-step imputation and control workflow. However, IDQN - IDQN method in Remedy does not outperform better than the original naive methods. Since, in this approach, all agents on \Imiss are only trained with imputed data which could bring in large uncertainty. We also observe training under this approach is hard to converge. 

% The detailed results are shown in Table ~\ref{tab:overall performance}

% \usepackage{multirow}
% \usepackage{graphicx}

\noindent We perform experiments to investigate how different approaches perform under different missing rates. The results can be found in Table~\ref{tab:overall performance}. We have the following observations:

\noindent $\bullet$ Compared with \fixfix, optimizing IDQN agents at \Iobs can significantly reduce the average travel time. This validates the effectiveness of RL agents over pre-timed agents. %but as the number of unobserved intersections increases, the performance gets worse and downgrades to the FixedTime control approach as all intersections observations are missing.

\noindent$\bullet$ Compared with \idqnfix, \idqnn works worse even at the most ideal settings and cannot converge at higher missing rates, which proves optimizing IDQN agents with no imputation cannot solve the missing data problem.

\noindent $\bullet$ \idqnidqn method in Remedy 2 does not outperform the original naive method. This is because, in this approach, all agents on \Imiss are only trained with imputed data which could bring in large uncertainty. %. We also observe training under this approach is hard to converge. 

\noindent $\bullet$ \sdqnsdqn (model-based), \sdqnsdqn (all), \sdqnsdqn (transferred), and \idqnmaxp approaches achieve better performances than naive \idqnfix approach under all three datasets. This proves the effectiveness of our two-step imputation and control method. As the missing rate increases, their performance decreases. %But with a missing rate greater than 20\%, only \sdqnsdqn (all) can achieve consistent improvement.
For shared-parameter methods, when the missing rate is moderate, the overall performance is not greatly affected. This is because the shared agent can learn from the experience in both $\mathcal{N}_o$ and $\mathcal{N}_m$ and make policy stable and easy to converge. 
More detailed intersection-level metrics can be found in Appendix~\ref{append: lane delay}.

\subsection{Data Sparsity Analysis}
In this section, we investigate how different missing rate and the location of unobserved intersections influences the performance of the proposed methods. 

%extend our experiment setting with neighboring and non-neighboring unobserved intersections to investigate how the proposed control approaches perform under these settings.

\vspace{1mm}
\noindent$\bullet$  \textbf{Influence of different missing rates.}
We randomly sample 1,2,3,4 intersections from 16 intersections in the $D_{HZ}$ and $D_{SYN}$ and 3,6,9,12 intersections from 48 intersections in the $D_{NY}$ as unobserved intersections. % and eliminate cases with neighboring unobserved intersections. 
The results in Table~\ref{tab:overall performance} and Figure~\ref{fig:non-neigbboring} show that when there are no neighboring unobserved intersections, \idqnmaxp, \sdqnsdqn (transferred), and \sdqnsdqn (all), \sdqnsdqn (model-based) achieve consistent better performances than \fixfix. The performances of the four approaches decrease as the number of unobserved intersections increases. Moreover, the three shared-parameter methods are more stable in performance improvement when the missing rates increase.

\vspace{1mm}
\noindent$\bullet$  \textbf{Influence of unobserved locations.}
In the previous experiments, the unobserved intersections are not adjacent to each other. Here we investigate how the locations of unobserved intersections influence the performance. We conduct experiments on situations where adjacent intersections are unobserved. 
We randomly sample missing intersections and make sure the network has two unobserved intersections adjacent.
%Since the imputation method is highly dependent on the upstream intersections, missing at neighboring intersections could negatively affect most control methods' performance in our two steps imputation framework. Thus 
% We randomly sample 2,3,4 intersections from 16 intersections in $D_{HZ}$ and $D_{SYS}$ and 6,9,12 intersections from 48 intersections in $D_{NY}$ and make sure at least two unobserved intersections are adjacent.
The result is shown in Figure~\ref{fig:neigbboring}. We have the following observations: (1) When there are adjacent unobserved intersections, our proposed method still outperforms the \fixfix method in most cases. Specifically, \sdqnsdqn (transferred), \sdqnsdqn (all), and \sdqnsdqn (model-based) perform consistently better than other baseline methods. (2) Except for \idqnfix, the performance of all other methods drops from non-neighboring scenarios to neighboring scenarios. This is likely because the performance of the control method relies on the imputation method, and missing data at neighboring intersections could negatively affect the performance of the imputation. %The details of the imputation error can be found in Appendix~\ref{sec: accuracy and performance}.

\begin{figure}[tb]
\centering
  \begin{tabular}{cc}
  \includegraphics[width=.2\linewidth]{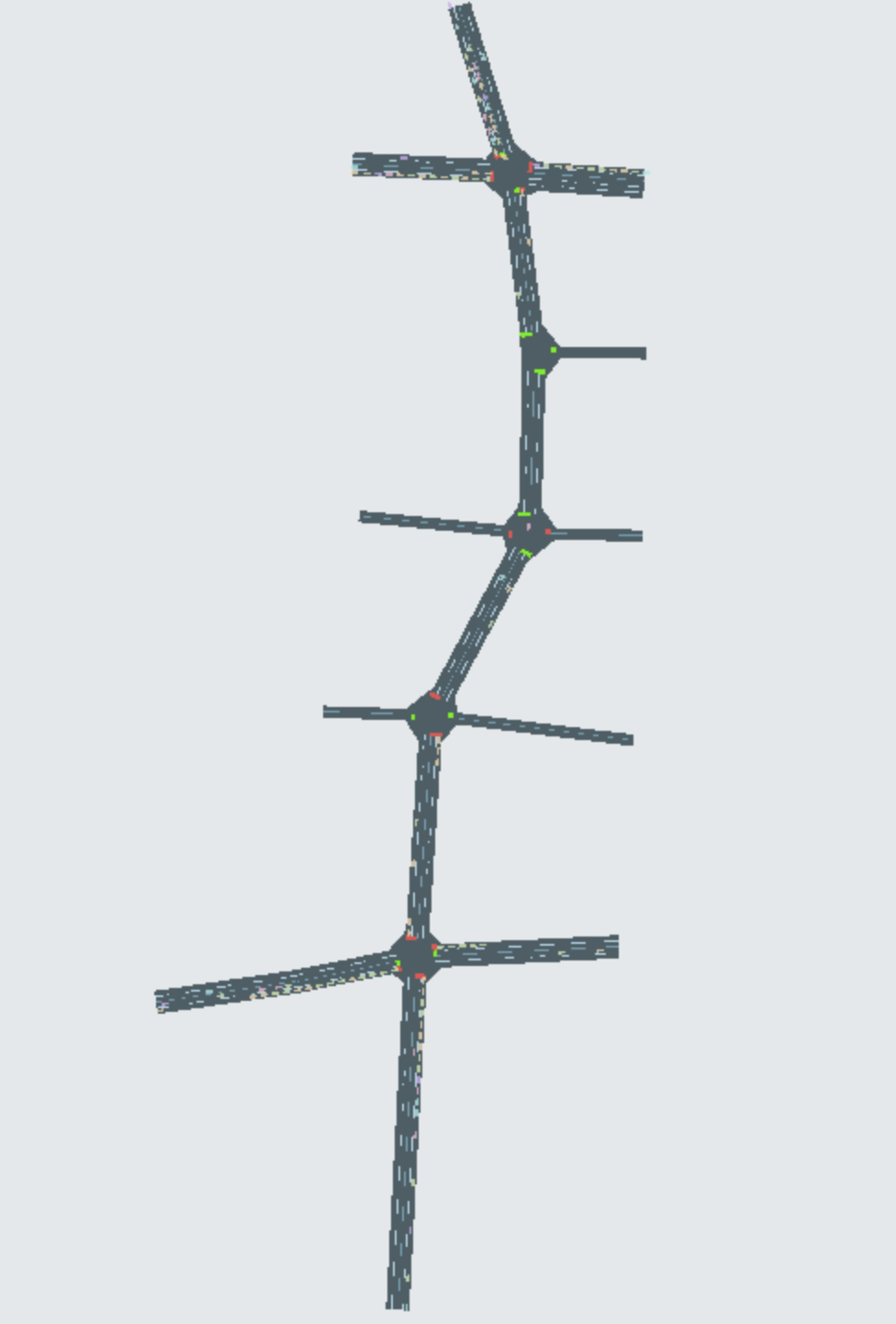} &
  \includegraphics[width=.44\linewidth]{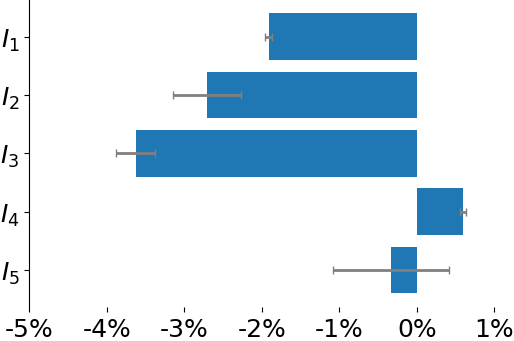} \\
  % \vspace{-1mm}
  (a) $D_{AL}$ road network & (b) Performance on $D_{AL}$ \\
  \end{tabular}
 \vspace{-2mm}
 \caption{(a) The topology of $D_{AL}$ network. (b) The decrease of averaged lane delays at each intersection between \fixfix and \sdqnsdqn (transferred) under one missing position. The more negative, the better. Other methods are shown in Appendix Figure~\ref{fig: Atlanta one}}
 \vspace{-3mm}
    \label{fig: heterogeneous}
\end{figure}

\vspace{0mm}
We also investigated the influence of missing data at frequently visited intersections and found no obvious performance drop when busy intersections are unobserved. The detailed results can be found in Appendix~\ref{sec: frequent visiting intersections}.
\subsection{Extension on Heterogeneous Intersections}
\label{sec: heterogeneous discussion}
In this section, we investigate our methods under heterogeneous intersections, $D_{AL}$, to see the applicability of our proposed method under the real-world setting. $D_{AL}$ is a public traffic dataset collected in Atlanta, GA, within $1\times 5$ intersections.
The intersections have different numbers of lanes and phases, leading to different state spaces for each agent. 
%The number of (lane, phase) pairs of each interaction in $D_{AL}$ are (6,2),(5,1),(6,2),(8,4), and (8,4), respectively.
% The states at each intersection are {6,5,6,8,8}, and action spaces are {2,1,2,4,4}.
We use FRAP ~\cite{DBLP:journals/corr/abs-1905-04722}, an RL model proven effective on heterogeneous structure road networks, as our base RL model for \sdqnsdqn to cope with the different input dimensions of heterogeneous intersections.
In this experiment, multiple runs are conducted, where each run has a random intersection selected as the unobserved intersection. 
% The result is shown in Fig~\ref{fig: heterogeneous}. 
% Compared to the pre-timed method, both proposed methods achieved considerable improvement. \sdqnsdqn (transferred) outperforms all other methods, which support that our two-step imputation method could help alleviate sparse observation problem even at highly heterogeneous structure datasets. 
We find the shared agent outperforms the baseline \fixfix method and reduces overall average travel time from 986.36 seconds (\fixfix) to 838.45 seconds. The detailed results in Figure~\ref{fig: heterogeneous} (b) show intersection-level average travel time in most intersections is reduced with our proposed method.

\section{Conclusion and Future Work}

In this paper, we investigate the traffic signal control problem in a real-world setting where the traffic condition around certain locations is missing. To tackle the missing data challenge, we propose two solutions: the first solution is to impute the state at missing intersections and directly use them to help agents at missing intersections make decisions; the second solution is to impute both state and reward and use the imputed experiences to train agents. We conduct extensive experiments using synthetic and real-world data and demonstrate the superior performance of our proposed methods over conventional methods. In addition, we show in-depth case studies and observations to understand how missing data influences the final control performance.

We would also like to point out several future works. First, missing data at neighboring intersections brings a considerable challenge to the imputation model. Future work could explore different imputation methods to improve imputation accuracy. Another direction is to combine our imputation model with more external data like speed data to make the imputation more accurate and improve control performance.
%Another direction is to combine the imputation with RL in one model, where currently, we only investigated a two-step approach which first imputes missing data and then trains the RL model. As discussed in the experiments, the uncertainty introduced by imputation makes RL algorithms hard to converge. Therefore, it is worthwhile to investigate how to improve the imputation accuracy and the RL algorithms' robustness to noisy samples in one model.

\clearpage
\bibliographystyle{named} 
\bibliography{ijcai23}
\clearpage

\appendix
\label{appendix}
\section{Appendix}

\arxiv{\subsection{Summary of Different Approaches}
We analyze and summarize all approaches in the Table~\ref{tab: pro-con},
 Generally, fixed-time methods (e.g., Fix-Fix, IDQN-Fix) do not rely on observations and cannot adapt well to dynamic traffic; shared RL methods (e.g., SDQN-SDQN) using Centralized Training and Decentralized Execution (CTDE) usually perform better than individual RL (e.g., IDQN), particularly with more agents, as validated by~\cite{chen2020toward}. SDQN-SDQN (transferred) in Remedy 1 uses control models only trained on observed intersections, which may underperform when deployed on unobserved intersections. SDQN-SDQN (all) in Remedy 2 pre-trains an additional reward model, enabling the update of the control model for unobserved intersections, thus mitigating performance issues. SDQN-SDQN (model-based) further refines the reward model training, enhancing the control model’s performance.

  \begin{table}[!ht]
  \caption{Summary of advantages and disadvantages of different approaches}
  \label{tab: pro-con}
\resizebox{\linewidth}{!}{

\setlength\extrarowheight{-6pt}
\begin{tabular}{|c|c|l|}
\hline
\textbf{Remedy} &
  \textbf{Approach} &
  \multicolumn{1}{c|}{\textbf{Advantage}} \\ \hline
\multirow{2}{*}{\textbf{Convention}} &
  \textbf{\fixfix} &
  \begin{tabular}[c]{@{}c@{}}
  The control model does not rely on imputation. \end{tabular}\\ \cline{2-3} 
 &
  \textbf{\idqnfix} &
  \begin{tabular}[l]{@{}l@{}}
  1. All the control models do not rely on imputation. \\
  2. IDQN is adaptive in observed intersections. \end{tabular}\\ \hline
\multirow{2}{*}{\textbf{Remedy 1}} &
  \textbf{\idqnmaxp} &
  \begin{tabular}[c]{@{}l@{}}
  1. Remedy 1 approaches enable adaptive control in \\ unobserved intersections. \\
  2. Unlike Remedy 2, Remedy 1 approaches do not \\ require a  reward imputation model.\end{tabular} 
  \\ \cline{2-3} 
 &
  \textbf{\begin{tabular}[c]{@{}c@{}}\sdqnsdqn\\ (transferred)\end{tabular}} &
  \begin{tabular}[c]{@{}l@{}}
  1. Remedy 1 approaches enable adaptive control in \\ unobserved intersections. \\
  2. Unlike Remedy 2, Remedy 1 approaches do not \\ require a reward imputation model.\\ 
  3. SDQN uses CTDE and converges  faster  in training \\ the control models.\end{tabular} \\ \hline
\multirow{3}{*}{\textbf{Remedy 2}} &
  \textbf{\idqnidqn} &
  \begin{tabular}[c]{@{}l@{}}Remedy 2 approaches enable the training of the \\ control model on unobserved intersections.\end{tabular} \\ \cline{2-3} 
 &
  \textbf{\begin{tabular}[c]{@{}c@{}}\sdqnsdqn\\ (all)\end{tabular}} &
  \begin{tabular}[c]{@{}l@{}}
  1. Remedy 2 approaches enable the training of the \\ control model on unobserved intersections. \\
  2. SDQN uses CTDE and converges faster in \\ training the control models.\end{tabular} \\ \cline{2-3} 
 &
  \textbf{\begin{tabular}[c]{@{}c@{}}\sdqnsdqn\\ (model-based)\end{tabular}} &
  \begin{tabular}[c]{@{}l@{}}
  1. Remedy 2 approaches enable the training of the \\ control model on unobserved intersections. \\
  2. SDQN uses CTDE and converges faster in \\ training the control models.  \\
  3. This method can alleviate the data-shifting \\ problem by updating the reward model.\end{tabular} \\ \hline
\textbf{} &
  \textbf{} &
  \multicolumn{1}{c|}{\textbf{Disadvatage}} \\ \hline
\multirow{2}{*}{\textbf{Convention}} &
  \textbf{\fixfix} &
  \begin{tabular}[c]{@{}l@{}}Fixed-time agents are not adaptive to dynamic traffic.\end{tabular} \\ \cline{2-3} 
 &
  \textbf{\idqnfix} &
  \begin{tabular}[c]{@{}l@{}}Fixed-time agents are not adaptive to dynamic traffic.\end{tabular} \\ \hline
\multirow{2}{*}{\textbf{Remedy 1}} &
  \textbf{\idqnmaxp} &
  \begin{tabular}[c]{@{}l@{}}
  1. Remedy 1 approaches require additional training \\ on the state imputation model. \\
  2. MaxPrusure is rule-based and has limited \\ performance  under complex traffic. \\
  3. IDQN is hard to converge with multiple agents.
  \end{tabular} \\ \cline{2-3} 
 &
  \textbf{\begin{tabular}[c]{@{}c@{}}\sdqnsdqn\\ (transferred)\end{tabular}} &
  \begin{tabular}[c]{@{}l@{}}
  1. Remedy 1 approaches require additional training \\ on the state imputation model. \\
  2. Control models are not trained on imputed \\ unobserved intersections.\end{tabular} \\ \hline
\multirow{3}{*}{\textbf{Remedy 2}} &
  \textbf{\idqnidqn} &
  \begin{tabular}[c]{@{}l@{}}
  1. Remedy 2 approaches require additional training \\ on reward imputation models.\\
  2. IDQN is hard to converge with multiple agents. \\
  3. Pretrained reward models might be biased \\ between observed and unobserved intersections.\end{tabular} \\ \cline{2-3} 
 &
  \textbf{\begin{tabular}[c]{@{}c@{}}\sdqnsdqn\\ (all)\end{tabular}} &
  \begin{tabular}[c]{@{}l@{}}
  1. Remedy 2 approaches require additional \\ training on reward imputation models.\\
  2. Pretrained reward models might be biased \\ between observed and unobserved intersections.\end{tabular} \\ \cline{2-3} 
 &
  \textbf{\begin{tabular}[c]{@{}c@{}}\sdqnsdqn\\ (model-based)\end{tabular}} &
  \begin{tabular}[c]{@{}l@{}} Remedy 2 approaches require additional training \\ on reward imputation models.\end{tabular} \\ \hline
\end{tabular}}
\end{table}
}

\subsection{Comparison of Different Imputation Models}
\label{tab: imputation model}
\begin{figure}[!ht]
    \centering
    \includegraphics[width=0.8\linewidth]{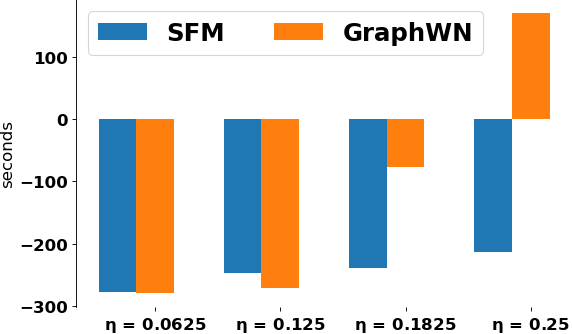}
    \caption{The decrease of two imputation models under different missing rates over \fixfix (conventional 1), w.r.t. the average travel time. The lower, the better}
    \label{fig:sfm_vs_graphwn.png}
\end{figure}

To investigate how different imputation models work with our proposed methods, we compare the SFM model with GraphWN, originally used for traffic forecasting problems, and use \sdqnsdqn (transferred) as the control method. Then we conduct experiments under different missing rates to explore different imputation models' performance on $D_{HZ}$. The result in Fig~\ref{fig:sfm_vs_graphwn.png} shows that at low missing rates, two imputation models achieve better performance than \fixfix (conventional 1). With the missing rate increases, the two imputation model's performance drops.  We also find the SFM model's performance degrades slower than the GraphWN method, though GrpahWN achieves better performance at lower missing rates. This shows that compared to learning neural networks, the rule-based SFM model performs more stable under high missing rates. And neural networks are potentially more effective, while they need more complete data.

\subsection{Performance of Different Control Methods on Individual Intersection}
\label{append: lane delay}
\begin{figure}[htb]
    
    \centering
    \subfigure[\idqnfix]{
    \includegraphics[width=0.45\linewidth]{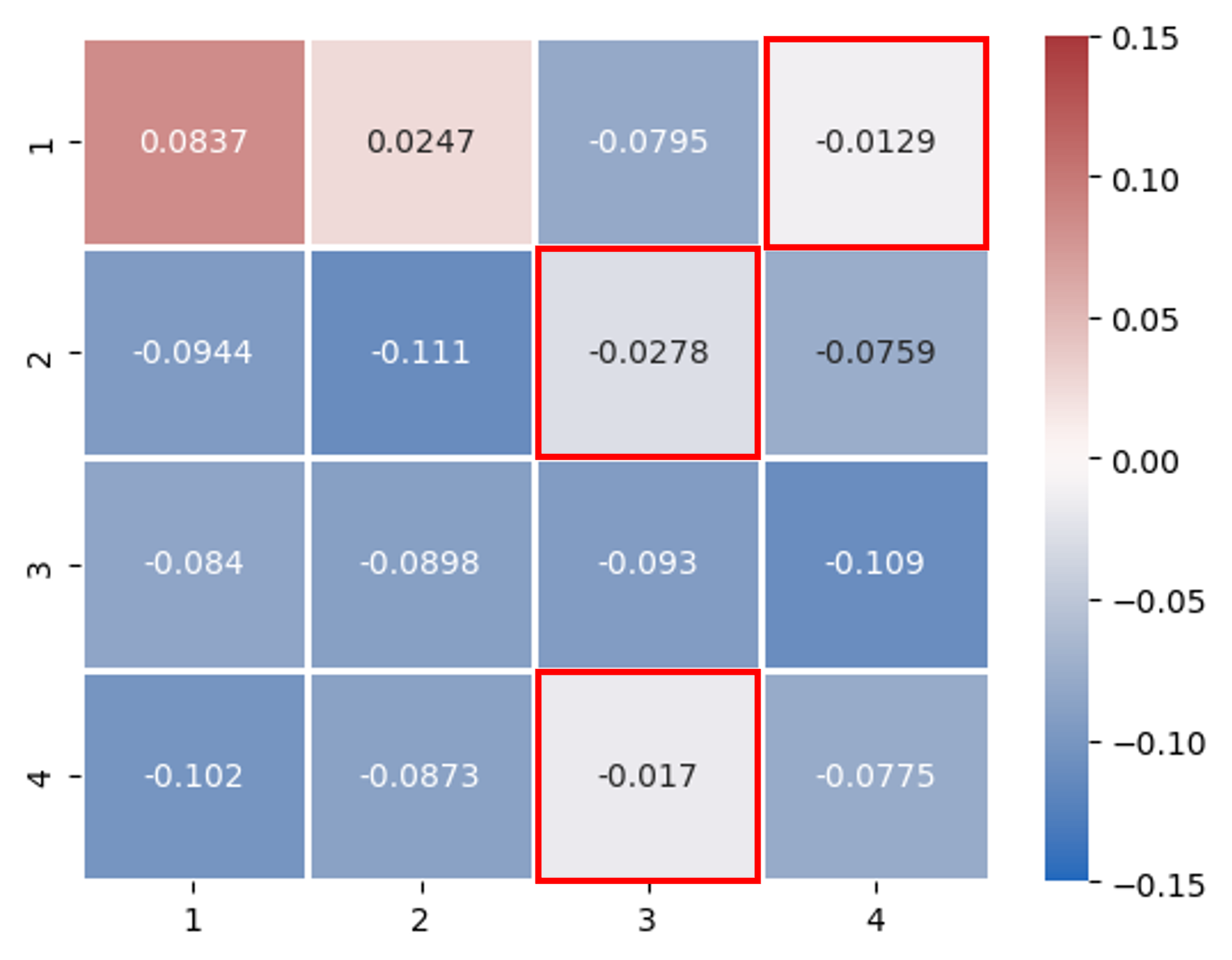}
    \label{tab: i-f delay}
    }
    \quad
    \subfigure[\idqnmaxp]{
    \includegraphics[width=0.45\linewidth]{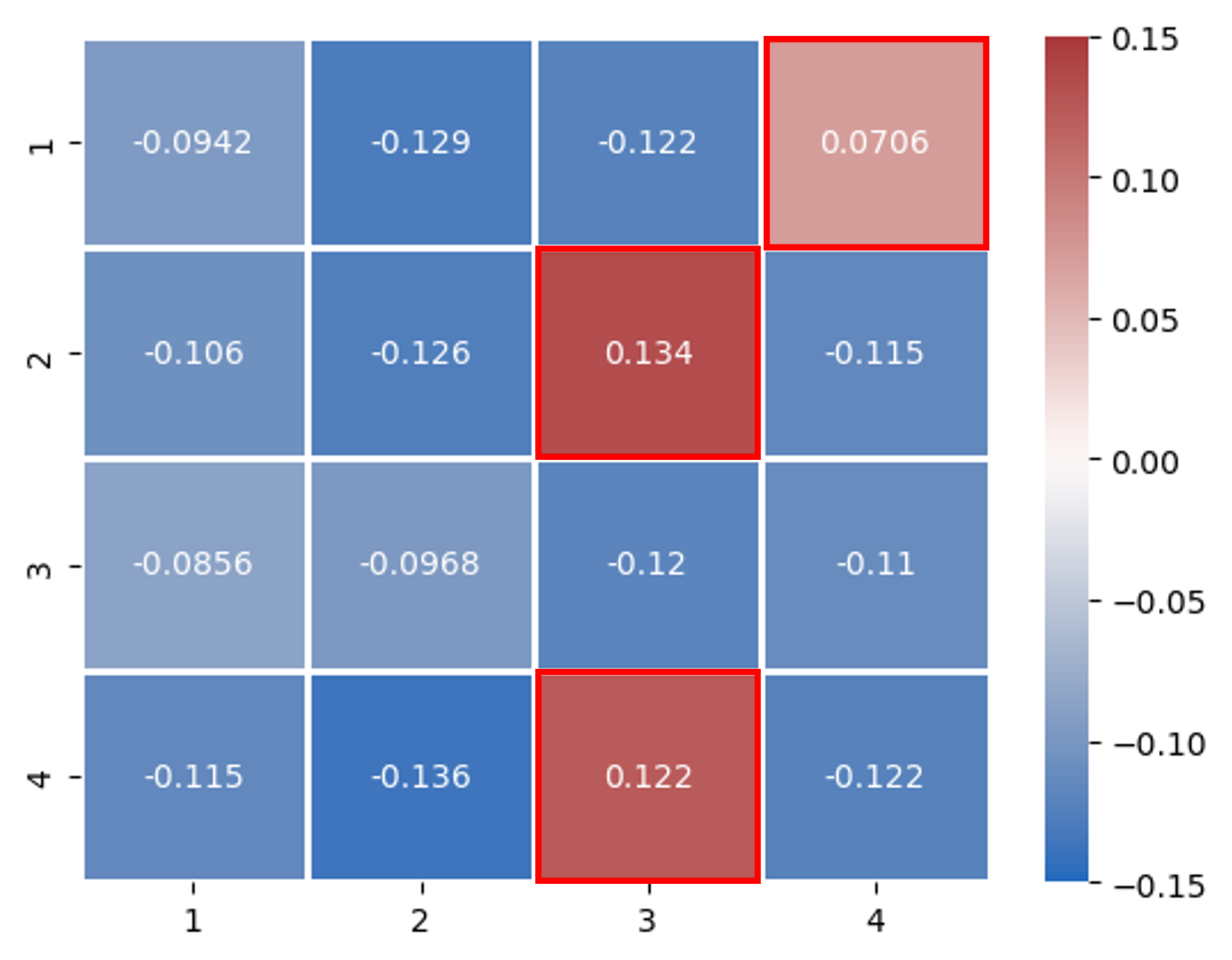}
    \label{tab: f-m delay}
    }
    \quad
    \subfigure[\sdqnsdqn (transferred)]{
    \includegraphics[width=0.45\linewidth]{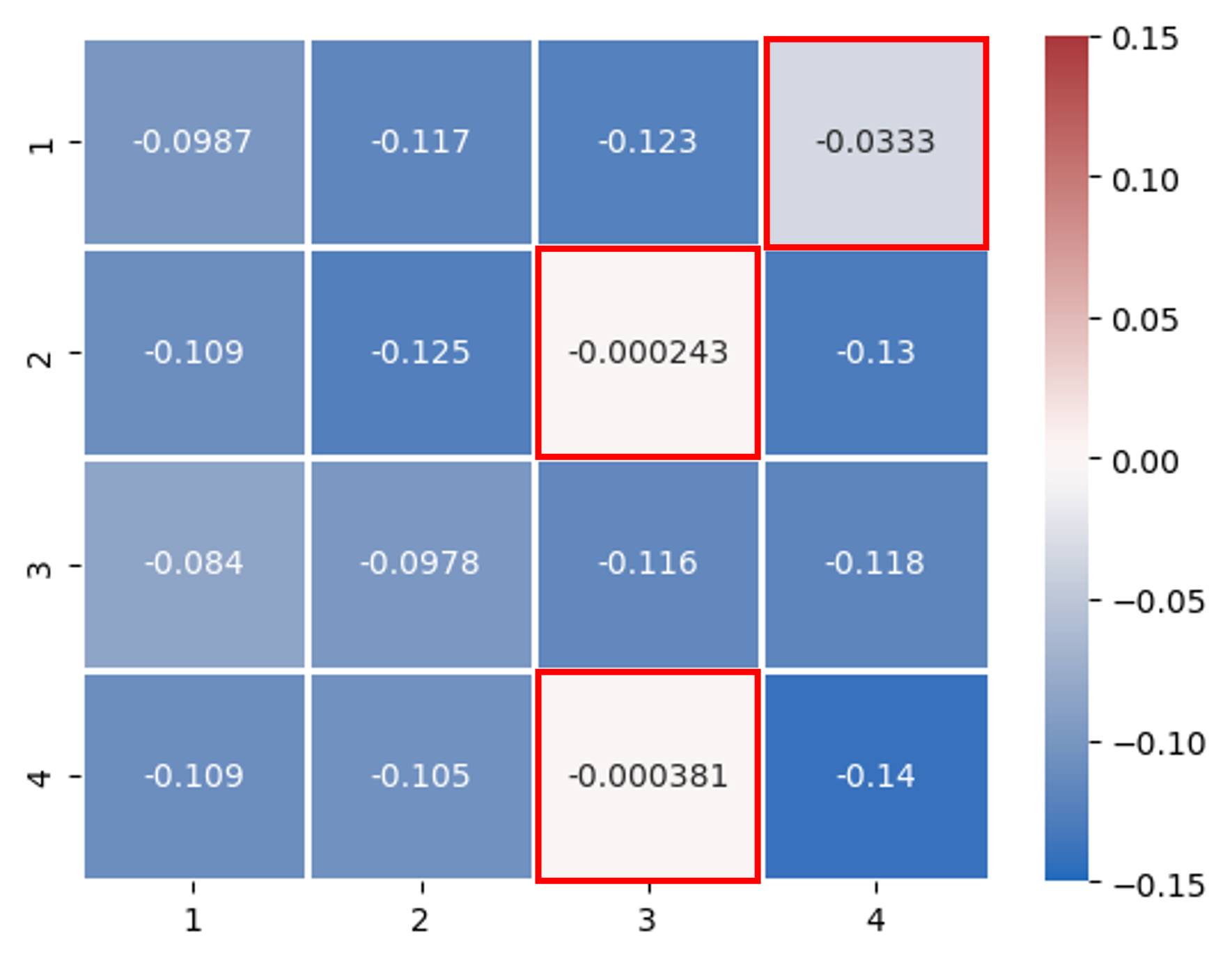}
    \label{tab: s-s-o delay}
    }
    \quad
    \subfigure[\idqnidqn]{
    \includegraphics[width=0.45\linewidth]{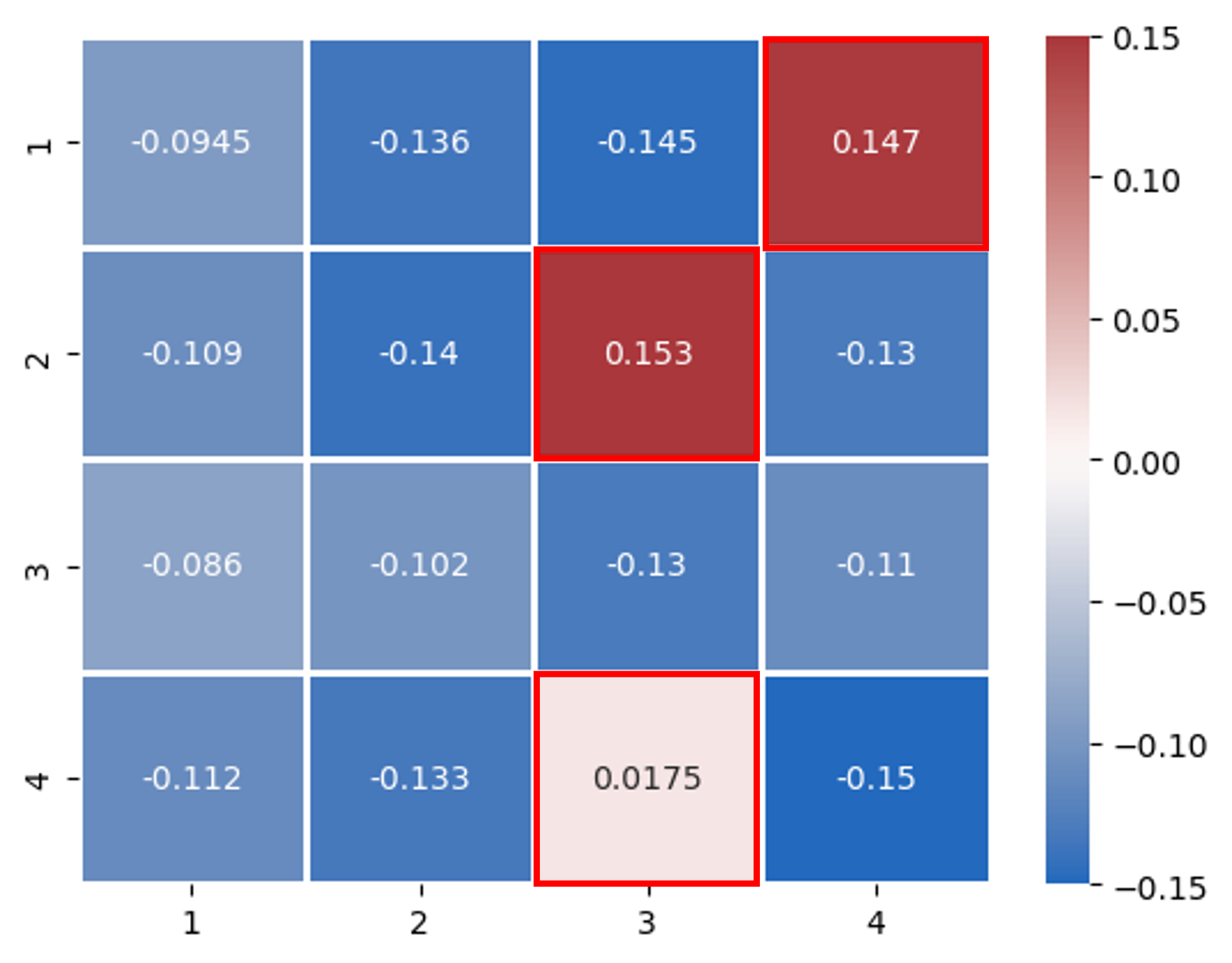}
    \label{tab:i-i delay}
    }
    \quad
    \subfigure[\sdqnsdqn (all)]{
    \includegraphics[width=0.45\linewidth]{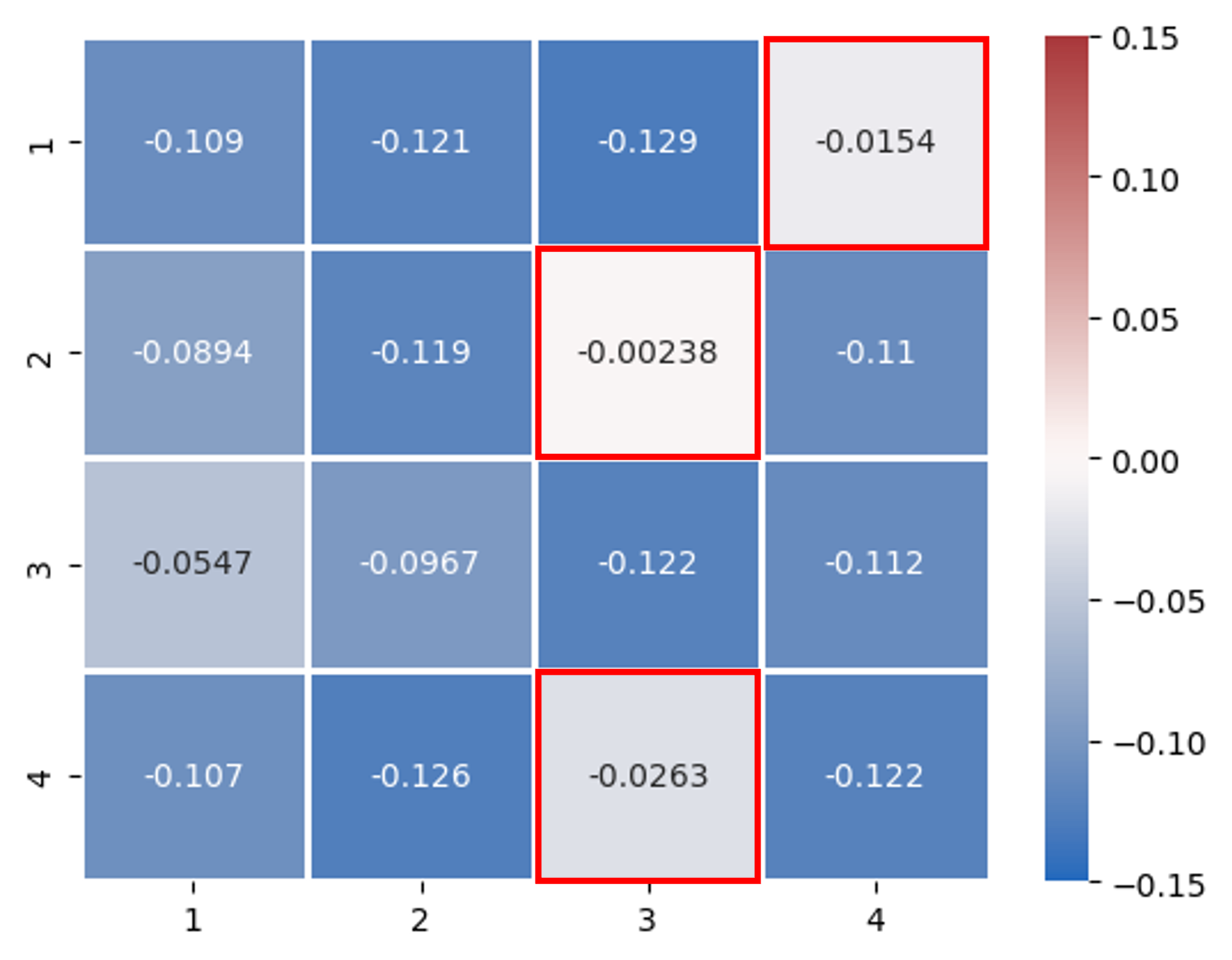}
    \label{tab: s-s-a delay}
    }
    \quad
    \subfigure[\sdqnsdqn(model)]{
    \includegraphics[width=0.45\linewidth]{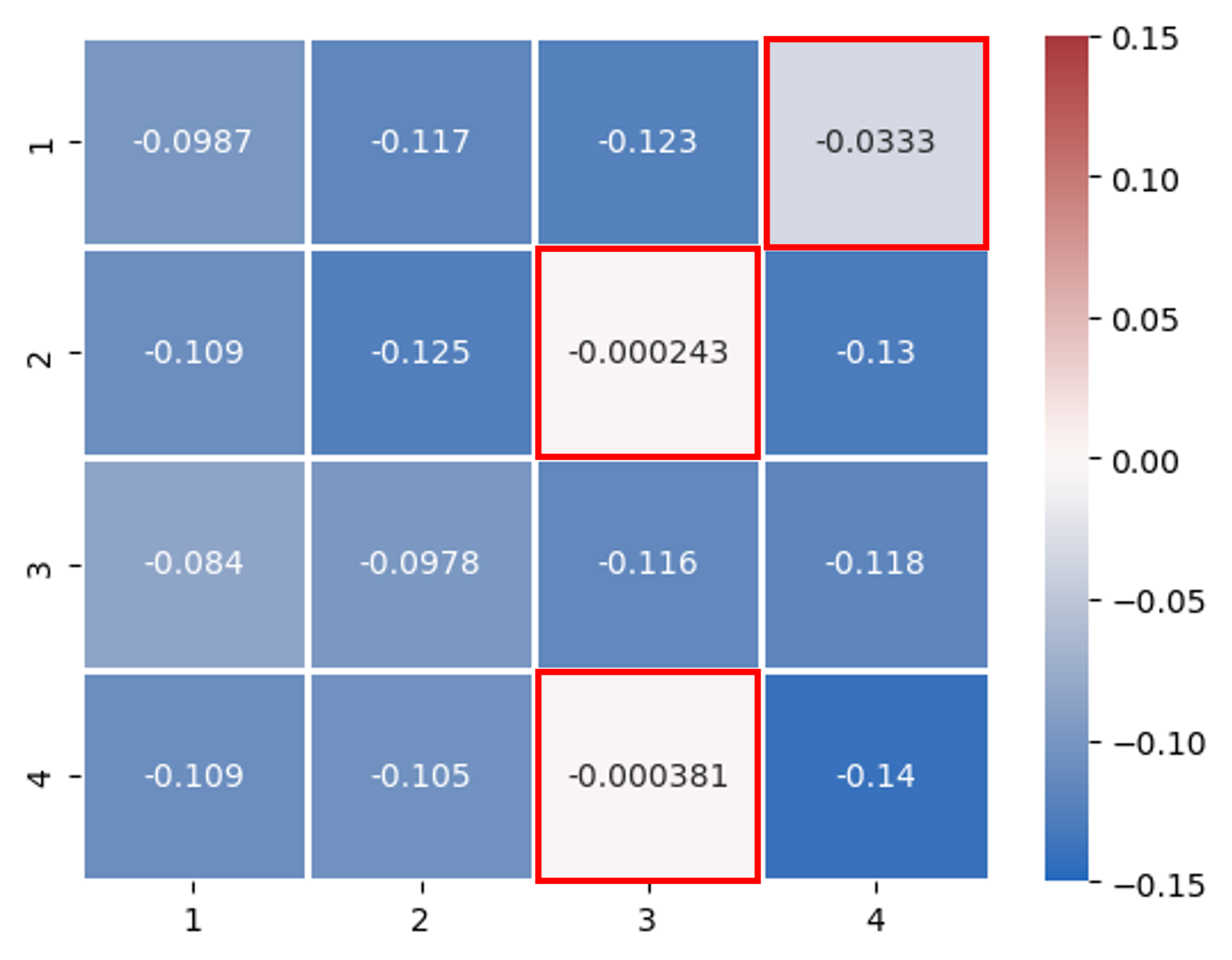}
    \label{tab: s-s-model-based delay}
    }
    \caption{The decrease of delay for six control approaches over \fixfix control method (conventional 1) with unobserved intersections at (1,4), (2,3), (4,3) marked with red rectangle. The more negative, the better performance compared to \fixfix control at each intersection.}
    \label{fig: lane delay}
\end{figure}

To explore how different control methods perform at the intersection level, we also report lane delay at each intersection and investigate different methods' influences. We randomly sample (1,4), (2,3), and (4,3) out of 16 positions as \Imiss and report lane delay. The result is shown in Figure~\ref{fig: lane delay}. We find \idqnfix approach can reduce delay at all \Iobs, but the optimization is limited. \idqnmaxp and \idqnidqn can significantly reduce lane delay at \Iobs. However, these two approaches also cause negative effects on \Imiss. This may be because of the imputation error at \Iobs, which significantly impairs RL-based and rule-based agents' decisions. For three approaches with parameter-sharing agents, though agents at \Iobs cannot achieve optimal value, compared to \fixfix and \idqnfix approaches, delay at most intersections, including \Imiss are reduced. 

\subsection{Influences of Unobserved Intersections at Frequently Visited Positions}
\begin{figure}[!ht]
    \centering
    \includegraphics[width=0.5\linewidth]{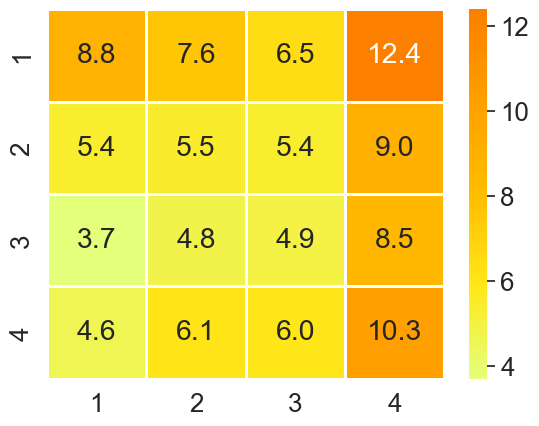}
    \caption{Total number of vehicles visited at each intersection (hundreds) in $D_{HZ}$. The higher number of visited vehicles means this intersection is busier and more important in the traffic.}
    \label{fig: visited freq}
\end{figure}

In traffic signal control, averages can hide adverse effects on sparsely visited intersections. We thus delve into a real-world dataset and investigate the effect of frequently visited intersections is unobserved. The total visited vehicle counts are shown in Figure~\ref{fig: visited freq}. 

\vspace{1mm}
\noindent$\bullet$  \textbf{Missing positions at frequently visited areas.}
\label{sec: frequent visiting intersections}
We sampled 3 groups of intersections, each has 3 out of 16 unobserved intersections, and tested different approaches' performance. The result is shown in Table~\ref{tab: frequently visited}. Comparing the average travel time of different approaches under the first two groups of intersections (not frequently visited) and the last group of intersections (frequently visited), we found no significant deprecation. 

The Last group includes one most visited intersections; the other two do not. Across the comparison between different imputation approaches, we find our methods are robust to this challenge. 

\begin{table}[htb]
\renewcommand{\arraystretch}{2}
\centering
\caption{Unobserved intersections allocated at different frequent vehicle visited positions. \underline{Frequently visited intersections} are highlighted }
\label{tab: frequently visited}
\resizebox{\columnwidth}{!}{%
\begin{tabular}{ccccccc}
\hline
\vspace{1mm}
MISSING POSITION &
  \idqnfix &
  \idqnmaxp &
  \begin{tabular}[c]{@{}l@{}}\sdqnsdqn\\ (transferred)\end{tabular} &
  \idqnidqn &
  \begin{tabular}[c]{@{}l@{}}\sdqnsdqn\\ (all)\end{tabular} &
  \begin{tabular}[c]{@{}l@{}}\sdqnsdqn\\ (model-based)\end{tabular} \\ \hline
(2,1), (2,4), (4,1)  & 367.43 & 344.24 & 331.53 & 457.86 & 332.76 & 330.53  \\
(1,1), ( 3,1), (3,4) & 369.47 & 343.42 & 338.07 & 581.97 & 330.18 & 336.05 \\
\underline{(1,4)}, (2,3), (4,3)  & 373.63 & 340.04 & 326.98 & 573.73 & 327.88  & 409.04 \\ \hline
\end{tabular}%
}
\end{table}

\subsection{Heterogeneous Road Network Topology}
The dataset $D_{AL}$ we used in Sec.~\ref{sec: heterogeneous discussion}
has a highly heterogeneous topology. It has 3 different configurations in total. Intersections 1 and 5 have eight lanes and four phases. Intersection 2 has five lanes and one phase, and intersections 3 and 4 have six lanes and two phases. 

\begin{figure}[!ht]
    \centering
    \includegraphics[width=\linewidth]{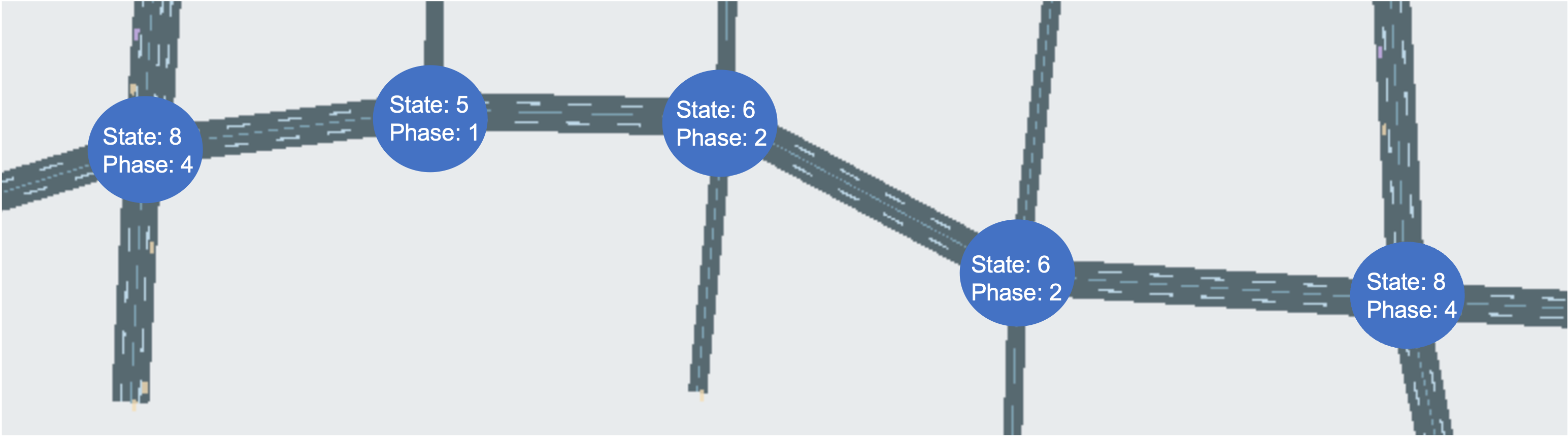}
    \caption{Topology of $D_{AL}$. It has 5 intersections with different states and phases. Indexes are 1 to 5 from left to right.}
    \label{fig: Atlanta details}
\end{figure}

\begin{figure}[!ht]
    \centering
    \includegraphics[width=0.8\linewidth]{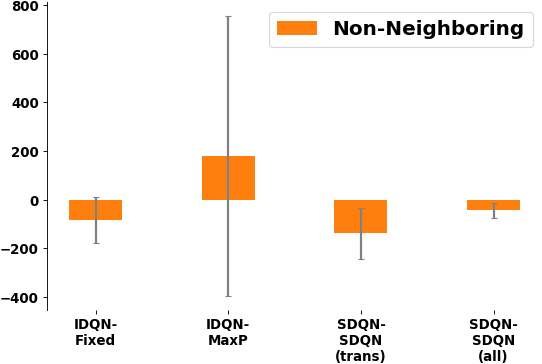}
    \caption{The decrease of average travel time for four control approaches over the \fixfix control method (conventional 1) with missing data
at one intersection. The more negative the value is, the better. \idqnfix, \sdqnsdqn (transferred), and \sdqnsdqn(all) perform better than \fixfix. And \sdqnsdqn (transferred) performs best among all}
    \label{fig: Atlanta one}
\end{figure}
We test our proposed methods on this heterogeneous structure network. The complete result of average travel time is shown in Figure~\ref{fig: Atlanta one} as support for the lane level result shown in Sec.\ref{sec: heterogeneous discussion}.
We found \sdqnsdqn (transferred) control method performs best, which supports the effectiveness of our proposed methods. Also, with parameter-sharing, our proposed two-step imputation methods are applicable to complex real-world scenarios.

\subsection{Base Model Analysis}
\label{sec:base-model}
We use DQN~\cite{wei2018intellilight} as our base model in the previous sections. To investigate the influences of the base model on the final performance, we conducted experiments using different RL models (i.e., FRAP~\cite{DBLP:journals/corr/abs-1905-04722}, Dueling DQN~\cite{DBLP:journals/corr/abs-2006-16712,DBLP:journals/corr/abs-2106-13174}) under the dataset $D_{HZ}$.

Table~\ref{tab:delta avg traveling} shows the performance comparison of the change on the base models with the DQN model w.r.t. average travel time. It can be observed that Dueling DQN can achieve better results than DQN in most cases. For other base models that are reported to be better than DQN, it is expected to improve the performance of our proposed method as well.

% \newpage
%chart and figure here
\begin{table}[htb]
  \centering
  \caption{$\Delta$ average travel time means reduced average travel time of FRAP and Dueling-DQN agents compared with DQN on $D_{HZ}$. \textbf{Negative} results show that with FRAP and Dueling DQN as RL agents, average travel time is reduced compared to DQN agents. \textbf{We highlighted results with $\Delta$ average travel time less than zero.}}
  \label{tab:delta avg traveling}
  \resizebox{\columnwidth}{!}{%
    \begin{tabular}{cccc}
\\  \toprule
    \textbf{METHOD}                                              & \textbf{MISSING RATE}             & \multicolumn{2}{c}%{$\mathbf{\Delta}$ \textbf{Average Travel Time}} \\ %\hline
{\textbf{$\Delta$ Average Travel Time}} \\ &  & \textbf{FRAP}          & \textbf{Dueling DQN}                 \\ 
                                                    \midrule
\idqnfix                                               & \multirow{6}{*}{6.25\%}  &        +32.0157       &      \textbf{-4.9868}              \\
\idqnmaxp                     &                          &           +53.7521          &       +11.1077       \\  
\sdqnsdqn (transferred)               &                                &          +25.736               &      \textbf{-1.6961}      \\
\idqnidqn                                               &                          &             \textbf{-22.8162}               &       \textbf{-67.5349}         \\
\sdqnsdqn (all)                                               &                          &           +22.4184          &        \textbf{-0.1957}        \\
\sdqnsdqn (model-based)                                                &                          &          +25.381           &        \textbf{-1.7308}        \\ \hline
\idqnfix                                               & \multirow{6}{*}{12.5\%}  &        +8.1247       &       \textbf{-16.7985}          \\
\idqnmaxp                                               &                          &      +23.5246         &      \textbf{-53.4985}              \\
\sdqnsdqn (transferred)                                               &                          &       +31.198           &               +17.7471        \\
\idqnidqn                                               &                          &             +60.4042              &          \textbf{-42.4859}      \\
\sdqnsdqn (all)                                              &                          &          +10.3246              &         +13.1825       \\
\sdqnsdqn (model-based)                                               &                          &         +7.532            &        +1.2769        \\ \hline
\idqnfix                                               & \multirow{6}{*}{18.75\%} &       +12.0974         &     \textbf{-14.1713}    \\
\idqnmaxp                                               &                          &        +45.5611       &         \textbf{-11.9744}           \\
\sdqnsdqn (transferred)                                              &                          &        \textbf{-10.5883 }          &      \textbf{-14.7092}          \\
\idqnidqn                                               &                          &               \textbf{-16.7969}           &        \textbf{-55.5823}        \\
\sdqnsdqn (all)  &                          &           \textbf{-5.1507}                &       \textbf{-0.4848}         \\
\sdqnsdqn (model-based)                                              &                          &        +66.0515           &     \textbf{-10.573}           \\ \hline
\idqnfix                                               & \multirow{6}{*}{25\%}    &       \textbf{-14.1858 }         &         \textbf{-31.6266}            \\
\idqnmaxp                                              &                          &     +107.4302          &       +6.8059        \\
\sdqnsdqn (transferred)                                               &                          &             +43.8644               &         \textbf{-12.0765}       \\
\idqnidqn                                               &                          &             +49.406           &    \textbf{-78.5265}            \\
\sdqnsdqn (all)                                               &                          &             \textbf{-6.9524}          &        \textbf{-9.9602}        \\
\sdqnsdqn (model-based)                                               &                          &           \textbf{-12.8496}          &       \textbf{-26.5348}         \\ \bottomrule
\end{tabular}}
\end{table}%

\end{document}